\documentclass[letterpaper,twocolumn,10pt]{article}
\usepackage{usenix2019_v3.1}

\usepackage{tikz}
\usepackage{amsmath}
\usepackage{amssymb}
\usepackage{array}
\usepackage{graphicx}
\usepackage{subcaption}
\usepackage{multirow}
\usepackage[ruled,vlined,linesnumbered]{algorithm2e}
\usepackage{booktabs,caption}
\usepackage{threeparttable}
\usepackage{color,soul}

\newtheorem{definition}{Definition}

\newcommand\MyBox[2]{
  \fbox{\lower0.75cm
    \vbox to 1.7cm{\vfil
      \hbox to 1.7cm{\hfil\parbox{1.4cm}{#1\\#2}\hfil}
      \vfil}%
  }%
}

\DeclareMathOperator{\E}{\mathbb{E}}

\begin{document}
%-------------------------------------------------------------------------------

%don't want date printed
\date{}

% make title bold and 14 pt font (Latex default is non-bold, 16 pt)
\title{Adaptive ABAC Policy Learning: A Reinforcement Learning Approach}

%for single author (just remove % characters)
\author{
{\rm Leila Karimi}\\
School of Computing and Information \\
University of Pittsburgh, Pittsburgh, USA\\
% {\tt\small lek86@pitt.edu}}
\and
{\rm Mai Abdelhakim}\\
Swanson School of Engineering \\
University of Pittsburgh, Pittsburgh, USA\\
% {\tt\small lek86@pitt.edu}}
\and
{\rm James Joshi}\\
School of Computing and Information \\
University of Pittsburgh, Pittsburgh, USA\\
% {\tt\small jjoshi@pitt.edu}}
}
\maketitle

%-------------------------------------------------------------------------------
\begin{abstract}
%-------------------------------------------------------------------------------
With rapid advances in computing systems, there is an increasing demand for more effective and efficient access control (AC) approaches. Recently,  Attribute Based Access Control (ABAC) approaches have been shown to be promising in fulfilling the AC needs of such emerging complex computing environments. An ABAC model grants access to a requester based on attributes of entities in a system and an authorization policy; however, its generality and flexibility come with a higher cost. Further, increasing complexities of organizational systems and the need for federated accesses to their resources make the task of AC enforcement and management much more challenging.
In this paper, we propose an adaptive ABAC policy learning approach to automate the authorization management task. We model ABAC policy learning as a reinforcement learning problem. In particular, we propose a contextual bandit system, in which an authorization engine adapts an ABAC model through a feedback control loop; it relies on interacting with users/administrators of the system to receive their feedback that assists the model in making authorization decisions. We propose four methods for initializing the learning model and a planning approach based on attribute value hierarchy to accelerate the learning process.  
We focus on developing an adaptive ABAC policy learning model for a home IoT environment as a running example. We evaluate our proposed approach over real and synthetic data. We consider both complete and sparse datasets in our evaluations. Our experimental results show that the proposed approach achieves performance that is comparable to ones based on supervised learning in many scenarios and even outperforms them in several situations. 

\textbf{Keywords}: Attribute Based Access Control, Policy Learning, Reinforcement Learning, Dynamic Authorization Policy, Usable Security
\end{abstract}

%-------------------------------------------------------------------------------
\section{Introduction}
%-------------------------------------------------------------------------------

With rapid advances in computing and information systems, technologies, and services (e.g. online social networks (OSNs), Internet of Things (IoT), cloud/edge computing, etc.), there is an increasing demand for more effective and efficient access control (AC) approaches to address the limitations of the existing ones. For decades, existing AC models such as Discretionary Access Control (DAC) \cite{sandhu1994access} \cite{harrison1976protection}, Mandatory Access Control (MAC) \cite{bell1973secure} \cite{sandhu1993lattice}, and Role-Based Access Control (RBAC) \cite{sandhu1996role} have played a significant role as key approaches in protecting information systems from unauthorized accesses to their resources. However, with increasing complexities of computing environments, such AC models have been shown to be inadequate in providing flexible, usable, and comprehensive authorization services \cite{hu2013guide}. For example, a health care environment requires a more comprehensive AC model that meets the needs of all parties involved in the health care ecosystem including patients, health care providers, and other stakeholders \cite{jin2009patient, karimi2017multi}, while considering the characteristics of various medical devices, sensors, and other contextual information.

Among several newer approaches proposed in the literature, Attribute Based Access Control (ABAC) has become quite popular as it has been shown to be a promising approach in addressing the authorization needs of emerging complex systems and environments \cite{hu2013guide}. An ABAC model grants access to a requester based on attributes of entities in the system, such as subject attributes, object attributes, environmental conditions, etc., and a set of authorization rules. Although ABAC has been shown to be superior to other existing models in many respects, its generality and flexibility come at a higher cost. ABAC systems can be much more complex than other AC models. Furthermore, increasing complexities of organizational systems and the need for federated accesses to their resources make adoption of ABAC and related AC management tasks much more challenging. In particular, organizations and information systems face the following challenges while employing an ABAC model:
\begin{itemize}
    \item Developing ABAC policy rules can lead to \emph{Rule Explosion}, as a system with $n$ number of attributes will have $2^n$ policy rule combinations. Hence, the manual ABAC policy development and management tasks are error-prone and tedious and a significant challenge even for medium sized organizations.
    \item In many systems (e.g., Home IoT environment), typically, users are responsible for defining authorization policies even though they may not be knowledgeable about nor capable of defining such complex policy rules. Hence, they may come up with policy rules that are not efficient, effective or correct/non-ambiguous.
    \item Modern information systems evolve rapidly, from the addition of new attributes and attribute values to the updates in authorization rules. Such dynamicity makes the management of ABAC policies even more challenging.
\end{itemize}

The above limitations are the primary motives behind the proposed adaptive ABAC policy learning framework to fully/partially automate the policy development and management task.

Recent research efforts have been focused on exploring Artificial Intelligence (AI) and Machine Learning (ML) based approaches for developing and managing ABAC authorization systems, ranging from mining ABAC policies from access logs \cite{karimi2018unsupervised, iyer2018mining, karimi2020automatic} to extracting such policies using deep learning (DL) algorithms \cite{mocanu2015towards} and employing natural language processing (NLP) tools for automating policy development and enforcement \cite{alohaly2019towards}. While supervised learning algorithms seem to be an option for inferring ABAC policies from access logs, they suffer from several limitations. First of all, supervised learning algorithms, especially DL algorithms, require a huge amount of labeled data showing which access requests should be permitted and which ones should be denied. Acquiring such labeled data is time-consuming and expensive and in some situations may not even be possible. Second, emerging and future computing and information systems are dynamic and highly interconnected. The authorization needs of the users and the attributes of the entities in the environment evolve rapidly. A supervised learning approach is incapable of adjusting to such dynamic settings or it needs a new set of labeled data to restart the learning process. Last but not the least, in most situations, the available access logs are often sparse and contain partial activity logs, and hence lack information about all possible access requests and the authorization decisions of the system for them. As a result, an ABAC model learned by a supervised learning algorithm over such data performs poorly in a real, constantly evolving system when encountering a novel access request. We believe that an adaptive access control approach that learns authorization rules from feedback provided by the users is a promising solution. Such an authorization framework shows promise in providing a usable and effective access control solution for complex environments. Reinforcement Learning (RL) provides a promising infrastructure for such adaptive authorization.

RL has become an active area in AI/ML, recently. In RL, agents learn to make better decisions by interacting with the environment. An agent begins by knowing nothing or very little about the given task and learns through reinforcements, i.e. rewards received through feedback from the environment showing how well it is performing the task. Lately, the combination of RL with deep learning - i.e., deep reinforcement learning - has proved to be a promising approach in mastering human-level control policies in various tasks \cite{mnih2015human}.

An RL algorithm for ABAC policy mining overcomes the limitation of a supervised learning approach. It does not require a huge amount of pre-labeled data while adjusting to a dynamic environment very well, and by receiving feedback from users, it learns about unknown access requests very quickly.

 In this paper, we utilize RL and, more specifically, contextual bandit, to establish a mapping between access requests and the appropriate authorization decisions for such requests. The authorization engine (AE) is considered as an agent in our proposed framework and its authorization decision (permit or deny) is an action according to the state of the system. Attributes of entities involved in an access request as well as the contextual factors form the state of the system. The AE learns authorization policies by interacting with the environment without having explicit knowledge of the original access control policy rules. We propose four methods for initializing the learning model and a planning approach based on hierarchies over attribute values to accelerate the learning process. We develop the proposed adaptive ABAC policy learning model, using a home IoT environment as a running example.
 
 To the best of our knowledge, our proposed model is the first RL based ABAC policy adaptation method that can be used to infer ABAC policy rules in complex environments where traditional methods are not effective and efficient.
 
 The rest of the paper is organized as follows. In Section \ref{preliminaries}, we overview the Attribute Based Access Control, Reinforcement Learning, Home IoT, and its authorization framework. In Section \ref{proposed}, we propose the reinforcement learning based framework for adaptive attribute based authorization learning while focusing on Home IoT as a running example. In Section \ref{evaluation}, we evaluates the proposed approach. In Section \ref{relatedwork}, we discuss the related work. Section \ref{discussion}, discusses some considerations that need to be reviewed while employing the proposed model. Finally, we present our conclusions in Section \ref{conclusion}. 
 
 \begin{table*}
\centering
\caption{Notations} \label{tab:notations}
\begin{tabular}{ll} 
\toprule
    Notation & Definition \\
\midrule
    $U$, $O$, and $OP$ & Sets of users, objects, and operations \\
    $UA$, $OA$, and $EA$ & Sets of user attributes, object attributes, and environmental attributes \\
    $E = U \cup O$ & Set of all entities in the system \\
    $ATTR = UA \cup OA \cup EA$ & Set of all attributes in the system \\
    $V(attr)$ & Set of all valid values for $attr$ in the system \\
    $F = \{ \langle attr, v \rangle | \: attr \in ATTR$ and $v \in V(attr) \}$ & Attribute Filter \\
    $q = \langle u, o, op, ea \rangle$ & An access request where user $u$ is the requester requesting to perform \\
    & operation $op$ over object $o$ while environmental conditions $ea$ holds\\
    AE & Authorization Engine (i.e. RL agent) \\
    $l = \langle q, d\rangle$ & Authorization tuple including decision $d$ made by the AE for request $q$ \\
    $\mathcal{L}$ & Access log (i.e. set of authorization tuples) \\
    $\rho = \langle uaf, oaf, eaf, op, d\rangle $ & ABAC policy rule \\
    $\mathcal{P}$ & Set of all ABAC policy rules in the system \\
    $\pi_{ABAC} = \langle E, ATTR, OP, \mathcal{P} \rangle $ & ABAC policy \\
    $\mathcal{S}$, $\mathcal{A}$, and $\pi$ & RL state space, action space, and policy \\
    $s_t = [ua_t, oa_t, ea_t, op]$ & system state at time step $t$ \\
    $a_t$ and $r_t$ & chosen action and reward (feedback) at time step $t$ \\
    $TP_w$, $TN_w$, $FP_w$ and $FN_w$ & reward function items, \\ & showing agreement or disagreement between owner and agent decision \\
    $\lambda_{TP}$, $\lambda_{TN}$ , $\lambda_{FP}$ and $\lambda_{FN}$ & weight of reward function items \\
    $d_w$ and $d_{AE}$ & decision of owner of an object and AE for an access request, respectively \\
    $owner(o)$ & a function that returns owners of object $o$ \\
    $VH_{attr}$ & Attribute Value Hierarchy \\
    $get\_state(q_t)$ & a function that returns a state $s_t$ corresponding to an access request $q_t$ \\
    $get\_request(s)$ & a function that returns an access request $q_s$ corresponding to a state $s$ \\
\bottomrule
\end{tabular}
\end{table*}

\begin{figure*}[htbp]
    \centering\includegraphics[clip, trim=0.1cm 15cm 0.1cm 1cm, scale=0.7]{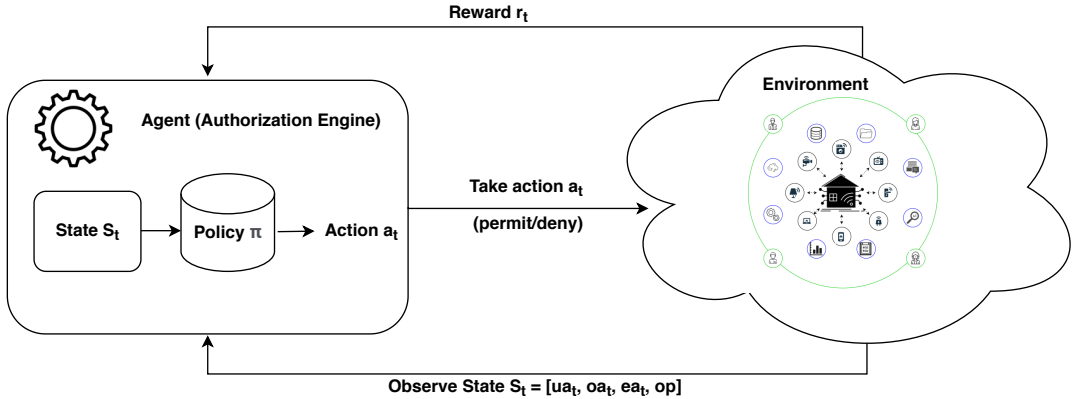}
    \caption{Adaptive Reinforcement Learning Based ABAC Policy Learning}
    \label{fig:framework}
\end{figure*}

%-------------------------------------------------------------------------------
\section{Preliminary} \label{preliminaries}
In this section, we overview Attribute Based Access Control (ABAC), Reinforcement Learning (RL), Home IoT, and its authorization framework.  Table \ref{tab:notations} summarizes the notations used in this paper.

\subsection{Attribute Based Access Control}
According to NIST's ``\textit{Guide to ABAC Definition and Consideration}" \cite{hu2013guide},  ``\textit{the ABAC engine can make an access control decision based on the assigned attributes of the requester, the assigned attributes of the object, environment conditions, and a set of policies that are specified in terms of those attributes and conditions}''. Throughout the paper, we use \emph{user attributes}, \emph{object attributes}, and \emph{environmental attributes} to refer to the attributes of the requester, attributes of the requested object, and the environmental conditions, respectively. Accordingly, the ABAC model has the following components:

$U$, $O$, and $OP$ represent sets of users, objects, and operations in a system, and $UA$, $OA$, and $EA$ correspond to sets of user attributes, object attributes, and environmental attributes, respectively. $E = U \cup O$ and $ATTR = UA \cup OA \cup EA$ are the sets of all entities and all attributes in the system.

\begin{definition}
(\textbf{Attribute Range}). 
Given an attribute $attr \in ATTR$, the \emph{attribute range} $V(attr)$ is the set of all valid values for $attr$ in the system.
\end{definition}

\begin{definition}
(\textbf{Attribute Filter}). 
An \emph{attribute filter} is defined as a set of tuples $F = \{ \langle attr, v \rangle | \: attr \in ATTR$ and $v \in V(attr) \}$. Here $\langle attr, v \rangle $ is an attribute filter tuple that indicates $attr$  has value $v$.
\end{definition}

\begin{definition}
(\textbf{Access Request}).
An \emph{access request} $q$ is a tuple $q = \langle u, o, op, ea \rangle$ where user $u \in U$ is the requester requesting to perform operation $op \in OP$ on object $o \in O$ while environmental attributes $ea \in EA$ holds.
\end{definition}

\begin{definition}
(\textbf{Authorization Tuple/Access Log}).
An \emph{authorization tuple} is a tuple $l = \langle q, d\rangle$ containing the final decision $d$ made by the authorization engine for request $q$. An \emph{Access Log} $\mathcal{L}$ is a set of such tuples.
\end{definition}

The decision $d$ of an authorization tuple can be \emph{permit} or \emph{deny}. The tuple with \emph{permit} decision means that user $u$ can perform operation $op$ over an object $o$ under environmental attributes $ea$. The authorization tuple with \emph{deny} decision means the user cannot get such access.

\begin{definition}
(\textbf{ABAC Rule}).
An \emph{ABAC rule} $\rho$ is a tuple $\rho = \langle uaf, oaf, eaf, op, d\rangle $, where $uaf$, $oaf$, and $eaf$ are user attribute filter, object attribute filter and environmental attribute filter, respectively, $op$ is a corresponding operation, and $d$ shows the decision of the ABAC rule for such combination of attributes and requested operation.
\end{definition}

\begin{definition}
(\textbf{ABAC Policy}).
An ABAC policy $\pi_{ABAC}$ is a tuple $\pi_{ABAC} = \langle E, ATTR, OP, \mathcal{P} \rangle $ where $E$, $ATT$, $OP$, and $\mathcal{P}$ are sets of entities, attributes, operations, and ABAC rules in the system, respectively.
\end{definition}
\subsection{Reinforcement Learning}
Reinforcement Learning (RL) refers to a set of algorithms that train an agent to make a sequence of decisions through an interaction with an unknown environment to attain a goal (i.e., maximize the expected cumulative discounted reward). The environment is modeled as a Markov decision process (MDP). At each time step $t$, the agent observes the current state $s_t$ of the environment from the state space $\mathcal{S}$. The agent takes an action $a_t$ from the action space $\mathcal{A}$ according to a policy $\pi$. Following the action, the agent receives a reward signal $r_t$, and the state of the environment transits to $s_{t+1}$. The goal of the agent is to attain a policy that maximizes the expected return $R$ which is the sum of future discounted rewards 

$$ R = \sum_{t=0}^{\infty}\gamma^tr_t$$

where $\gamma \in [0,1]$ is a discount rate that determines the significance of future rewards. 

Policy $\pi$ is a probability distribution over actions given the states,

$$\pi : \pi(s, a) \to [0,1].$$

Here, $\pi(s, a)$ is the likelihood of action $a$ in state $s$. For each MDP, there exists an optimal policy $\pi^*$ that is at least as good as all other policies, expressed as follows: 

$$\pi^* \geq \pi \qquad \forall \pi$$

The value function of policy $\pi$ in state $s$ is the expected total reward for an agent starting at state $s$:

$$V^\pi(s) = \E[R_t \mid s_t = s]$$

Similarly, the Q-function of policy $\pi$ is defined as the expected return from choosing action $a$ in state $s$, and the following policy $\pi$ afterward:

$$Q^\pi(s, a) = \E[R_t \mid s_t = s, a_t=a]$$

The optimal Q-function denotes the maximum reward we can expect by selecting action $a$ in state $s$:

$$Q^*(s, a) = \max_{\pi} Q^\pi(s, a)$$

Knowing the optimal Q-function, we can easily extract the optimal policy by selecting an action which results in the maximum $Q^*(s, a)$ for each state:

$$\pi^*(s) = \arg \max_{a} Q^*(s, a) \qquad \forall s \in \mathcal{S}$$

It is common to use a \textit{function approximator} to estimate the function Q, especially when there are many possible \{state, action\} pairs. A function approximator has a set of adjustable parameters, $\theta$, referred to as \textit{policy parameters}.  

%%%To Do: \textbf{TALK ABOUT DEEP NURAL NETWORK THAT WILL BE USED AS FUNCTION APPROXIMATOR}

\subsubsection{Contextual Bandit}
Contextual bandit is an extension of the n-armed bandit problem which is a simplified form of RL problems. A contextual bandit formulates a round-by-round interaction between a learner and an environment while introducing contextual information in the interaction loop. The learner uses the contextual information to select the best action in each round. Then the learner observes a loss for the chosen action only. Contextual bandit based approaches are suitable for various real-world interactive machine learning problems. As learner receives limited feedback from the environment, exploration plays an important role in contextual bandit algorithms.

%%%To Do: \textbf{TALK ABOUT THE REWARD THAT IS NOT DELAYED}

\subsection{Home IoT and Its Authorization Framework}
% Home IoT consists of several devices that are connected through the internet. These devices either produce data or have a capability (e.g. locking the door, playing the music, etc.) that is used by the users of the system. The data produced is consumed by various applications, is used by users of the system, or actuates other physical devices. Such a complex environment has various accesses and information flows that need to be protected:

% \begin{itemize}
%     \item Third-party applications accessing data produced and performing computation over it;
%     \item Users accessing or consuming the data produced for making various decisions;
%     \item Users accessing the capabilities of an IoT device to perform an operation on it (e.g turning on a lamp using their smartphone);
%     \item An IoT device output actuating other physical devices through some control operations that need to be authorized.
% \end{itemize}

% Throughout the paper, we focus on a home IoT as a running example and apply our proposed approach for adapting an access control model for it that authorizes users of the Home IoT to access capabilities of IoT devices.

IoT is one of the emerging paradigms that is evolving quickly. It allows users to be connected to various objects and exchange data across their networks. Home IoT is one example of such an environment where users in a household interact with internet-connected devices that include sensors and appliances in their daily routines. Home IoT devices are usually managed through a central controller that handles the communication between devices, enforces users' authorization policies, and often allows for the execution of programs and applications over such devices. ABAC model is a good fit as an authorization approach for such an environment. However, despite traditional information systems with single-user devices (e.g. computers, phones, etc.), in a home IoT environment, multiple users interact with each device. As a result, traditional authorization frameworks fail to provide usable and flexible access-control specification and enforcement in such settings. With numerous IoT devices, a wide range of attributes and contextual factors, and various relationships between users and devices, it is unthinkable to expect ordinary users to be able to manage such complex authorization requirements and enforcement infrastructure. In this paper, we focus on a home IoT as a running example and apply our proposed approach for adapting an access control model for it.

We focus on a capability-centric authorization model instead of a device-centric one as suggested by \cite{he2018rethinking} where a capability is defined as an operation (e.g., play music) that can be performed on an IoT device (e.g., Google Home). The capability-based authorization model is more suitable for home IoT, as each home IoT device may have several capabilities with different sensitivity levels that results in different access control policy rules for various capabilities of an IoT device. For example, assume that a smart door lock has various capabilities such as unlocking, locking, viewing the state of the lock, deleting logs, and so on. In this example, unlocking the door is more sensitive than viewing a state of the lock and hence needs a more restricted AC policy rule. 

% \begin{figure}[htbp]
%     \centering\includegraphics[clip, trim=2cm 10cm 2cm 1cm, scale=0.5]{Home_IoT.pdf}
%     \caption{Overview of Home IoT Environment.}
%     \label{fig:HomeIoT}
% \end{figure}

% As shown in Figure \ref{fig:HomeIoT},

A home IoT environment consists of a set of users interacting with a set of home IoT devices. Each IoT device has a set of capabilities and a set of owners/administrators who manage the accesses to such capabilities. Here, users, devices, and capabilities are equivalent to set of users $U$, set of objects $O$, and set of operations $OP$, respectively, in the corresponding ABAC authorization model.

When a user wants to access a capability of a device (e.g. the user wants to play music on Google Home), his access request is directed to the authorization engine (AE). The AE evaluates the access request and sends its authorization decision back to the system to be enforced. The authorization decision is either \textit{permit} or \textit{deny}, granting or preventing requested access to the user, respectively. The AE evaluates the access requests based on the states of the system and the policy it learned through reinforcement learning algorithm (detailed in Section \ref{proposed}).

%%%To Do: \textbf{Talk about over-assignments and under-assignments, and FP FN and rewards}

%-------------------------------------------------------------------------------
\section{Adaptive ABAC Policy Learning} \label{proposed}
In this section, we introduce the ABAC-RL, an adaptive reinforcement learning based ABAC policy learning framework. We present the key components of ABAC-RL in detail and propose several methods for improving the model.

\subsection{ABAC-RL Framework} \label{RL_framework}
Figure \ref{fig:framework} shows the overall ABAC-RL framework. Here, an agent corresponds to the authorization engine (AE) that adapts an ABAC model through a feedback control loop by interacting with the environment. The environment is considered to mainly have the users/administrators of the system. An agent begins by knowing nothing or very little about the authorization policies of the administrators of the system and learns through reinforcements (i.e. rewards/penalties received through feedback from the environment showing how well it is deciding on access requests). 

At each time step $t$ (i.e. when an access request $q_t$ is received by the system), the AE agent observes a state, $s_t$, from the state space, $\mathcal{S}$, and accordingly takes an action, $a_t$, from the action space, $\mathcal{A}$, determining authorization decision based on the policy, $\pi$. In our model, the state observed by an AE agent for characterizing the environment consists of several parts: attributes $ua_t$ of user $u_t$ requesting the access, attributes $oa_t$ of object $o_t$,  environmental attributes,$ea_t$, and operation $op_t$ that is requested to be performed on object $o_t$. Hence, the state can be shown as $s_t = [ua_t, oa_t, ea_t, op_t]$. Accordingly, the action (i.e., the authorization decision) is either \textit{permit} or \textit{deny}, granting or preventing requested access to the requester, respectively.

Following an action, an agent receives a reward, $r_t$, determined by the feedback from the users of the system showing their agreement/disagreement with the authorization decision of the AE agent. For this purpose, the authorization decision of AE for each request is recorded. The administrator of the requested object checks the records and submits his feedback to the system showing his agreement/disagreement with the authorization decision of AE. We note that administrators usually give feedback when they do not agree with the authorization decision of AE, otherwise they may not give any feedback. Hence, we assume no feedback from the administrator means he agreed with the decision.

The feedback provided by administrators of a requested object will be used in the reward function to calculate the reward of the action that was taken by an AE agent for the corresponding access request. The reinforcement learning algorithm will use the calculated reward to update its policy with the goal of making better decisions in the future. We explain the details of the reward function in the following section.

The action chosen by the RL agent would result in a loss. If the action chosen matches the decision of the administrator of the system for corresponding access request, the loss will be zero and if the action chosen does not match, the loss value for such a decision is 1. The loss rate of an agent for a period of time is the average of the loss values for all the decisions that have been made by the agent in that period. The administrator of system can set a desirable threshold for loss rate of the model and employ the model in real-world scenarios only when the loss rate of the model is less than the set threshold (more details in Section \ref{discussion}).

\begin{table*}
\centering
\begin{threeparttable}
    \caption{Reward Function Items Based on Agent and Owner decisions for an Access Request}
    \label{table:confusion}
    \centering
    \begin{tabular}{|p{6.5cm}|p{6.5cm}|} \hline
        \vspace{.25\baselineskip}
        $TP_w =
        \begin{cases}
          1 & \text{if $d_w = permit$ and $d_{AE} = permit$}\\
          0 & \text{otherwise}
        \end{cases}$
        \vspace{1mm}
        &
        \vspace{1mm}
        $FP_w =
        \begin{cases}
          1 & \text{if $d_w = deny$ and $d_{AE} = permit$}\\
          0 & \text{otherwise}
        \end{cases} $
        \vspace{1mm}
        \\ \hline
        \vspace{1mm}
        $FN_w =
        \begin{cases}
          1 & \text{if $d_w = permit$ and $d_{AE} = deny$}\\
          0 & \text{otherwise}
        \end{cases}$ 
        \vspace{2mm}
        &
        \vspace{1mm}
        $TN_w =
        \begin{cases}
          1 & \text{if $d_w = deny$ and $d_{AE} = deny$}\\
          0 & \text{otherwise}
        \end{cases} $ 
        \vspace{2mm}
        \\ \hline
    \end{tabular}
    \begin{tablenotes}
      \item $d_w$ and $d_{AE}$ represents decision of the owner of an object and the authorization agent for an access request, respectively.
    \end{tablenotes}
 \end{threeparttable}
\end{table*}

\subsection{Reward Function} 
At each time step and for each access request, the AE agent takes an action (i.e. permit or deny) according to the current state, $s_t \in \mathcal{S}$ and based on its policy $\pi$. The goal of the agent is to minimize unauthorized access to the objects. In order to reach this objective, the authorization decision of the agent should match the collective authorization decisions of the owners/administrators of the corresponding object. Therefore, the reward function is set up in a way to achieve this goal.
The input to the reward function is the feedback of all owners/administrators of the requested object. If the feedback shows an agreement with AE's decision, the agent should get a positive reward to learn that its decision was correct. On the other hand, when the feedback shows a disagreement between an owner and the decision of an AE, the agent should receive a negative reward so it will adjust its policy with the goal of making better decisions in the future. Hence, we propose a reward function as follows:
\begin{equation*}
  \begin{gathered}
$$ r_t = \\
\sum\limits_{w \in owner(o_t)}{\lambda_{TP} \cdot TP_w + \lambda_{TN} \cdot TN_w -\lambda_{FP} \cdot FP_w - \lambda_{FN} \cdot FN_w} $$
\end{gathered}
\end{equation*}

where, $owner(o_t)$ is a function that returns all owners (or administrators) of object $o_t$ that was requested at time step $t$, and $TP_w$, $TN_w$, $FP_w$ and $FN_w$ are reward function items that are calculated based on the decision of AE and the feedback from owner of the device (see details in Table \ref{table:confusion}), and $\lambda_{TP}$, $\lambda_{TN}$ , $\lambda_{FP}$ and $\lambda_{FN}$ are their corresponding weights. Here, true positives and true negatives are represented as positive rewards as the goal of the agent is to maximize them while both false positives and false negatives are represented as penalties, as the goal is to minimize these measures.

% \noindent
% \renewcommand\arraystretch{1.5}
% \setlength\tabcolsep{0pt}
% \begin{tabular}{c >{\bfseries}r @{\hspace{0.7em}}c @{\hspace{0.4em}}c @{\hspace{0.7em}}l}
%   \multirow{10}{*}{\rotatebox{90}{\parbox{3cm}{\bfseries\centering Owner Decision}}} & & \multicolumn{2}{c}{\bfseries Agent Decision} & \\
%   & & \bfseries permit & \bfseries deny & \bfseries \\
%   & permit & \MyBox{True}{Positive} & \MyBox{False}{Negative} & \\[2.4em]
%   & deny & \MyBox{False}{Positive} & \MyBox{True}{Negative} &
% \end{tabular}

\subsection{Policy Initialization Techniques}
Reinforcement learning algorithms start with the initialization of a target policy. We propose four different approaches for initializing an ABAC-RL policy in our framework. These initialization methods may overlap and the real-world system can employ any of these approaches or all of them, together. Here again, we use Home-IoT as our running example. 

\subsubsection{Initialization with General AC Policy Rules}
As studied by He \textit{et al.} in \cite{he2018rethinking}, a set of desired access control policies are typically consistent among IoT Home users. For example, it is desirable that all users be capable of controlling the lights and thermostats when they are at home. As another example, deleting the lock log should be denied for all users of the system except the owners. Multiple such candidate general policies have been suggested by He \textit{et al.} in  \cite{he2018rethinking}. These general AC policies are a good starting point for initializing the ABAC-RL policy.

\subsubsection{Initialization with Default User Settings}
At the beginning of employing an RL for learning authorization policies of a Home-IoT system or when a new user (e.g. a baby sitter) is added to the Home-IoT, the owners of the devices can define a few access control policy rules as a default setting of the authorization framework. These settings are employed by the RL algorithm to initialize the corresponding policies so it will converge to the desired authorization policies faster and prevent over-privileged or under-privileged accesses. For example, the default authorization policy rule for "neighbor" could be \emph{deny} while the default policy rule for "parent" could be \emph{permit}.

\subsubsection{Default Decision for Capabilities}
Different capabilities of Home IoT devices have different levels of sensitivity. For a highly sensitive capability, any over-privileged access could result in a serious loss of security or privacy and is against the \emph{principle of least privilege} \cite{saltzer1975protection}. The default decision (i.e., RL action) for an access request corresponding to such a capability should be \emph{deny}. On the other hand, for nonsensitive capabilities, an under-assignment could result in an authorized user being denied from accessing the capability which is very inconvenient. Such under-assignments could adversely affect the availability of an object/capability and should be avoided. Hence, the default decision for an access request corresponding to such a capability should be \emph{permit}. The default decisions for such capabilities can be set as part of the initialization of the authorization policy. For example, the default decision for "check\_temperature" is \emph{permit}.

\subsubsection{Initialization with Past Access Logs}
Information systems' owners often desire to employ modern access control models and migrate from outdated authorization models to new ones. Various policy learning methods have been proposed to automate such migration \cite{xu2015mining, cotrini2018mining, iyer2018mining, karimi2018unsupervised, karimi2020automatic}. In the case of policy learning, available access logs from former AC models can be used to form a new refined policy. In our proposed approach, we utilize the available access logs to initialize the corresponding RL model.

\subsection{Policy Learning with Planning}
In a reinforcement learning model, an agent improves its decision making strategy by interacting with the environment. In each state, the agent chooses an action and receives feedback for the chosen action. Based on the received feedback, it will update its policy for decision making in that state. By experiencing more and more state-action-feedback sequences, the agent policy will get close to the optimal policy. However, for a large space of states, the agent does not have complete information for all the states. Its information is partial as it may not visit all the states. The agent can improve its information on an unseen state by utilizing the information of neighboring states. 

In our ABAC-RL policy learning framework, we propose a planning strategy to enhance the learned policy in presence of partial information. Such a strategy will help the agent to make a better decision for previously unseen states. The proposed planning algorithm is based on pre-defined hierarchies of attribute values in the system. Attribute value hierarchy defines a quasi ordering between different values of an attribute in the model. Formally, we define an attribute value hierarchy as follows:

\begin{definition} \textbf{Attribute Value Hierarchy}
For each attribute $attr \in (UA \cup OA \cup EA)$, the attribute value hierarchy $VH_{attr} \subseteq V(attr) \times V(attr)$ is a partial order on $V(attr)$ called closeness relation, written as $v_1 \succeq v_2$,  where $v_1$ is called the upper value and $v_2$ is called the lower value in the relation.
\end{definition}

The proposed planning strategy identifies neighboring states and conclude the access decision of one state based on another. To this purpose, we formally define the neighboring states as follows:

\begin{definition} \textbf{Neighboring States}
Two states $s_{t_1}$ and $s_{t_2}$ are called neighboring states where for all attribute $attr \in (UA \cup OA \cup EA)$, except $attr'$ the two states have the same value, $attr' = v_1$ in state $s_{t_1}$ and  $attr' = v_2$ in state $s_{t_2}$, and $attr'$ has a value hierarchy where $v_1$ and $v_2$ are in closeness relationship (i.e. $v_1 \succeq v_2$ or $v_2 \succeq v_1$).
\end{definition}

Two \emph{Upper Neighboring State} and \emph{Lower Neighboring State} are defined as follows to distinguish between concluding "permit" and "deny" decisions for neighboring states. 

\begin{definition} \textbf{Upper and Lower Neighboring States}
Given two neighboring states $s_{t_1}$ and $s_{t_2}$ where the value of all their attributes except $attr'$ are the same and $attr' = v_1$ in state $s_{t_1}$ and  $attr' = v_2$ in state $s_{t_2}$, the state $s_{t_1}$ is called the upper neighboring state and state $s_{t_2}$ is called the lower neighboring state if $v_1$ and $v_2$ are in closeness relationship and $v_1 \succeq v_2$.
\end{definition}

% \begin{definition} \textbf{Attribute Value Hierarchy}
% For each attribute $attr \in (UA \cup OA \cup EA)$, the attribute value hierarchy $VH_{attr} \subseteq V(attr) \times V(attr)$ is a partial order on $V(attr)$ called neighborhood relation, written as $v_1 \succeq v_2$,  where permitting states $s_1$ concluded based on permitting state $s_2$ and denying state $s_2$ is concluded based on denying state $s_1$. Here, $V(attr)$ is a set of all possible values for attribute $attr$, $s_1$ and $s_2$ are two neighboring states where the values of all attributes except $attr$ are the same, and the values of $attr$ in state $s_1$ and $s_2$ are $v_1$ and $v_2$, respectively.
% \end{definition}

The intuition behind the planning strategy is that the authorization decision for a state is similar to the authorization decision for its neighboring state (with higher/lower order attribute value in the same hierarchy). In our planning process, when the AE receives feedback for an access log, it records the decision for the access log and all its neighboring states. For each access log, we only consider the first level neighboring states, meaning that we only consider the unseen states that differ in one attribute value with the given state. Algorithm \ref{algo:planning} shows the details of the planning process.

As an example assume that "minor" and "teenager" are two possible values of attribute "age\_range" in Home-IoT. There is a closeness relation between these two attribute values written as $teenager \succeq minor$ meaning that states $s_{teenager}$ and $s_{minor}$ are neighboring states and given that all other attribute values of these two states are equal, permitting an action in the state with $age\_range = teenager$ is expected if authorization decision for an action in the state including $age\_range = minor$ is \emph{permit}. So if an action is permitted for a minor child in a home IoT environment it will be permitted for a teenager in the same situation. On the other hand, we decide to deny an action in a state for a minor child if the action is denied for a teenager in the same circumstances.

\begin{algorithm}[!t]
\SetAlgoLined
\caption{Planning Algorithm}
\label{algo:planning}
\KwIn{$\mathcal{L}$}
\KwOut{$\mathcal{L}$}
\SetKwProg{Fn}{procedure}{}{}

\Fn{Planning}{
    \ForAll{$l=[q_t,d_t] \in \mathcal{L}$}{
        $s_t = get\_state(q_t)$\;
        \If{$d_t == ``permit"$}
        {
            $S\_Upper = get\_upper\_neighbors(s_t)$\;
            \ForAll{$s \in S\_Upper$}
            {
                \If{$s \notin \mathcal{L}$}
                {
                    $q_s = get\_request(s)$\;
                    $\mathcal{L} = \mathcal{L} \cup [q_s, ``permit"]$\;
                }
            }
            
        }
        \If{$d_t == ``deny"$}
        {
            $S\_Lower = get\_lower\_neighbors(s_t)$\;
            \ForAll{$s \in S\_Lower$}
            {
            \If{$s \notin \mathcal{L}$}
                {
                    $q_s = get\_request(s)$\;
                    $\mathcal{L} = \mathcal{L} \cup [q_s, ``deny"]$\;
                }
            }
            
        }
    }
    \Return $\mathcal{L}$
}
\end{algorithm}

\section{Evaluation}\label{evaluation}
We have implemented a prototype of our proposed approach presented in Section \ref{proposed}. In this section, we present our experimental evaluation.

\subsection{Experiment Setup}
Evaluation over real authorization data would be ideal, however, as we only have access to one real access log from Amazon, we developed various sample policies, attribute data, and their corresponding synthesized access logs. We developed two sets of sample policies, one with manually written policy rules and the other with randomly generated policy rules.  Each sample policy is the desired policy that the ABAC-RL algorithm's goal is to learn. We generate a synthesized access log for each sample policy. To generate the synthesized access log, we brute force through all attributes and their values to produce all possible combinations for the access tuples. We use this method to generate a complete access log for the manual and random policies. In the synthesized access logs, each access tuple corresponds to an access request and the desired authorization decision based on the original policy. To check the feasibility of our approach over sparse data, we also consider a partial dataset for each sample policy. We produce the sparse dataset (partial dataset) by randomly selecting access log records from the complete dataset. The ABAC-RL learning algorithm is run over each dataset to see how it will learn the authorization policy gradually.

The log generation and the proposed adaptive ABAC-RL learning algorithm are written in Python 3. We use
Vowpal Wabbit (VW) \footnote{https://vowpalwabbit.org} 
for implementing various contextual bandit methods (see details in Section \ref{baselines}) as well as a supervised learning algorithm. The supervised learning algorithm is a one-against-all logistic regression method provided by VW. The experiments were performed on a 64-bit Windows 10 machine having 8 GB RAM and an Intel Core i7 processor.

The performance of each method on a dataset with $n$ records is measured by the \textit{progressive
validation loss} (P.V.Loss) \cite{blum1999beating} that is calculated as follows:

$$PVL = \frac{1}{n} \sum_{t=1}^n{c_t a_t}$$

where $a_t$ is the loss for each chosen action ($a_t = 0$ if the chosen action matches the decision of the original access tuple and $a_t = 1$ otherwise) in time step $t$ and $c_t$ is the corresponding cost ($c_t = 1$ in our experiments). In our experiments, we assume each object has one owner and the weights of reward function items ($\lambda$s) equal 1.

% I a real-world situation, the agent executes the proposed planning algorithm for each access request after its authorization decision is finalized (i.e. after receiving feedback from the administrators). In our experiments, as we have all access logs beforehand, we apply the planning algorithm over each access tuple and add the corresponding access tuples of neighboring states after such tuple in the access log before training the RL algorithm.

\subsubsection{Datasets}
To evaluate the proposed model, we perform our experiments on multiple datasets including synthesized and real ones. The synthesized authorization records are generated based on two sets of ABAC policies: a set of ABAC policies with manually written policy rules and a one with randomly generated policy rules. The real dataset is built from the records provided by Amazon in Kaggle competition \cite{kaggle}.

% and Amazon Access data set from the UCI machine learning repository \cite{uci_amazon_access}.

\textbf{Real Dataset - Amazon Kaggle:}
The Kaggle competition dataset \cite{kaggle} includes access requests made by Amazon’s
employees in a two year period. Each access tuple in this dataset corresponds to an employee’s request to a resource and shows whether the access was permitted or not. The access log consists of the employees' attribute values and the resources' identifier. The dataset includes more than 12,000 users and 7,000 resources. %%%To Do: Work on this information

% \textbf{Real Dataset - Amazon UCI:}
% This dataset is provided by Amazon through the UCI machine learning repository \cite{uci_amazon_access}. It includes more than 36,000 users and 27,000 permissions. The dataset contains over 33000 attributes. %%%To Do: Work on this information

\textbf{Synthesized dataset - manually written policies:} We developed a set of sample policies including manually written rules and attribute data. We generated synthesized access log data for each manual policy. The access log consists of access requests and the desired decision based on the corresponding policy rules. Appendix \ref{sample} shows the details of the manually written policies. 

% To generate the synthesized access log, we brute force through all attributes and attribute values to produce all possible combinations of them. This procedure generates a complete access log for each sample policy. 

\textbf{Synthesized dataset - randomly generated policies:}
The authorization rules for these policies are generated completely randomly from random sets of attributes and attribute values. These randomly generated policies provide an opportunity to evaluate our proposed approach on datasets with various sizes and with varying structural characteristics. 

Table \ref{tab:policies_details} shows the details of the access log datasets. In this table, $|\mathcal{P}|$ shows the number of ABAC rules in the original policy, $|ATTR|$ and $|V| = \sum\limits_{attr \in ATTR}|V(attr)|$ show the number of attributes and attribute values, respectively, and $|\mathcal{L}|$ shows the size of access log.

\begin{table}
\centering
\caption{Details of Datasets} \label{tab:policies_details} 
  \begin{tabular}{lcccc}
    \toprule
    $\pi_{ABAC}$ & $|\mathcal{P}|$ & $|A|$ & $|V|$ & $|\mathcal{L}|$ \\
    \midrule
    Kaggle \cite{kaggle} & - & 9 & 15626 &  ~33K\\
    Manual Policy 1 ($\pi_{m1}$) & 11 & 5 & 30 & ~6K\\
    Manual Policy 2 ($\pi_{m2}$) & 11 & 5 & 29 & ~5K\\
    Manual Policy 3 ($\pi_{m3}$) & 38 & 5 & 44 & ~48K\\
    Synthetic Policy 1 ($\pi_{s1}$) & 5 & 8 & 30 & 21K \\
    Synthetic Policy 2 ($\pi_{s2}$) & 10 & 10 & 34 & 70K \\ 
    Synthetic Policy 3 ($\pi_{s3}$) & 15 & 12 & 37 & 200K \\
    \bottomrule
    \end{tabular}
\end{table}

\subsection{Contextual Bandit Algorithms} \label{baselines}
We empirically evaluated four contextual bandit algorithms for our proposed model. The algorithms are as follows:
\begin{itemize}
    \item $\epsilon$-greedy algorithm \cite{langford2008epoch}: The algorithm greedily exploits the best action learned with probability $1 - \epsilon$ and explore uniformly over all actions with probability $\epsilon$.
    \item Explore-first: The algorithm exclusively explores the first $k$ trials and then exploits the best action learned afterward. 
    \item Bagging: The algorithm trains multiple policies using bootstrapping. Given the context, the algorithm samples from the distributions over the actions provided by these policies.
    \item Online cover \cite{agarwal2014taming}: The algorithm explores all available actions while keeping only a small subset of policies active.
\end{itemize}

\begin{table*}
\small
\centering
\caption{Progressive validation loss, best hyperparameter values, and running times of various algorithm on different datasets}
\begin{tabular}{llcccccc}
    \hline
    \multirow{2}{*}{\textbf{Databases}}&&\multicolumn{6}{c}{\textbf{Algorithms}}\\\cline{3-8}
    &&$\epsilon$-greedy & Explore-first & Bagging & Online Cover & Planning & Supervised\\
    \hline
    \multirow{3}{*}{Kaggle \cite{kaggle}}& \textbf{P.V.Loss}& 0.065 & 0.058 & 0.059 & 0.058 & - & 0.055\\
    &\textbf{Best Hyperparameter}&$\epsilon=0.01$ & 10 first & 2 bags &cover $n = 2$ & - & NA\\
    &\textbf{Running Time (s)}&0.588&0.400&0.617&0.459&-&0.351\\
    \hline
    \multirow{3}{*}{Manual Policy 1 ($\pi_{m1}$)}& \textbf{P.V.Loss}&0.16&0.2&0.13&0.11&-&0.14\\
    &\textbf{Best Hyperparameter}&$\epsilon=0.01$ & 1500 first & 4 bags &cover $n = 2$ & - & NA\\
    &\textbf{Running Time (s)}&0.115&0.132&0.148&0.162&-&0.130\\
    \hline
    \multirow{3}{*}{Manual Policy 2 ($\pi_{m2}$)} & \textbf{P.V.Loss}&0.13&0.17&0.11&0.08&-&0.10\\
    &\textbf{Best Hyperparameter}&$\epsilon=0.02$ & 300 first & 2 bags &cover $n = 2$ & - & NA\\
    &\textbf{Running Time (s)}&0.095&0.132&0.093&0.121&-&0.139\\
    \hline
     \multirow{3}{*}{Manual Policy 3 ($\pi_{m3}$)} & \textbf{P.V.Loss}&0.07&0.1&0.04&0.03&0.02&0.05\\
    &\textbf{Best Hyperparameter}&$\epsilon=0.01$ & 10 first & 2 bags &cover $n = 2$ & cover $n = 2$ & NA\\
    &\textbf{Running Time (s)}&0.346&0.355&0.377&0.388&.401&0.301\\
    \hline
    \multirow{3}{*}{Synthetic Policy 1 ($\pi_{s1}$)} & \textbf{P.V.Loss}&0.15&0.14&0.09&0.08&0.07&0.12\\
    &\textbf{Best Hyperparameter}&$\epsilon=0.02$ & 1500 first & 10 bags & cover $n = 1$ & cover $n = 1$ & NA\\
    &\textbf{Running Time (s)}&0.203&0.275&0.303&0.217&.295&0.232\\
    \hline
    \multirow{3}{*}{Synthetic Policy 2 ($\pi_{s2}$)} & \textbf{P.V.Loss}&0.11&0.09&0.08&0.06&0.05&0.06\\
    &\textbf{Best Hyperparameter}&$\epsilon=0.03$ & 1500 first & 6 bags & cover $n = 1$ & cover $n = 1$ & NA\\
    &\textbf{Running Time (s)}&0.466&0.742&0.620&0.410&0.495&0.569\\
    \hline
    \multirow{3}{*}{Synthetic Policy 3 ($\pi_{s3}$)} & \textbf{P.V.Loss}&0.12&0.11&0.07&0.06&0.05&0.07\\
    &\textbf{Best Hyperparameter}&$\epsilon=0.01$ & 1500 first & 8 bags & cover $n = 1$ & cover $n = 1$ & NA\\
    &\textbf{Running Time (s)}&6.2&6.1&6.1&6.1&6.1&6.1\\
    \hline
\end{tabular}
\label{table:comparison}
\end{table*}

\subsection{Experimental Results}
In this section, we compare various contextual bandit algorithms for policy learning over different databases. We also evaluate our proposed ABAC-RL framework against several baselines. Table \ref{table:comparison} reports the result of the best parameter settings for each algorithm over different databases.

\subsubsection{Kaggle Access Control Dataset}
\begin{figure}[htbp]
    \centering\includegraphics[scale=0.4]{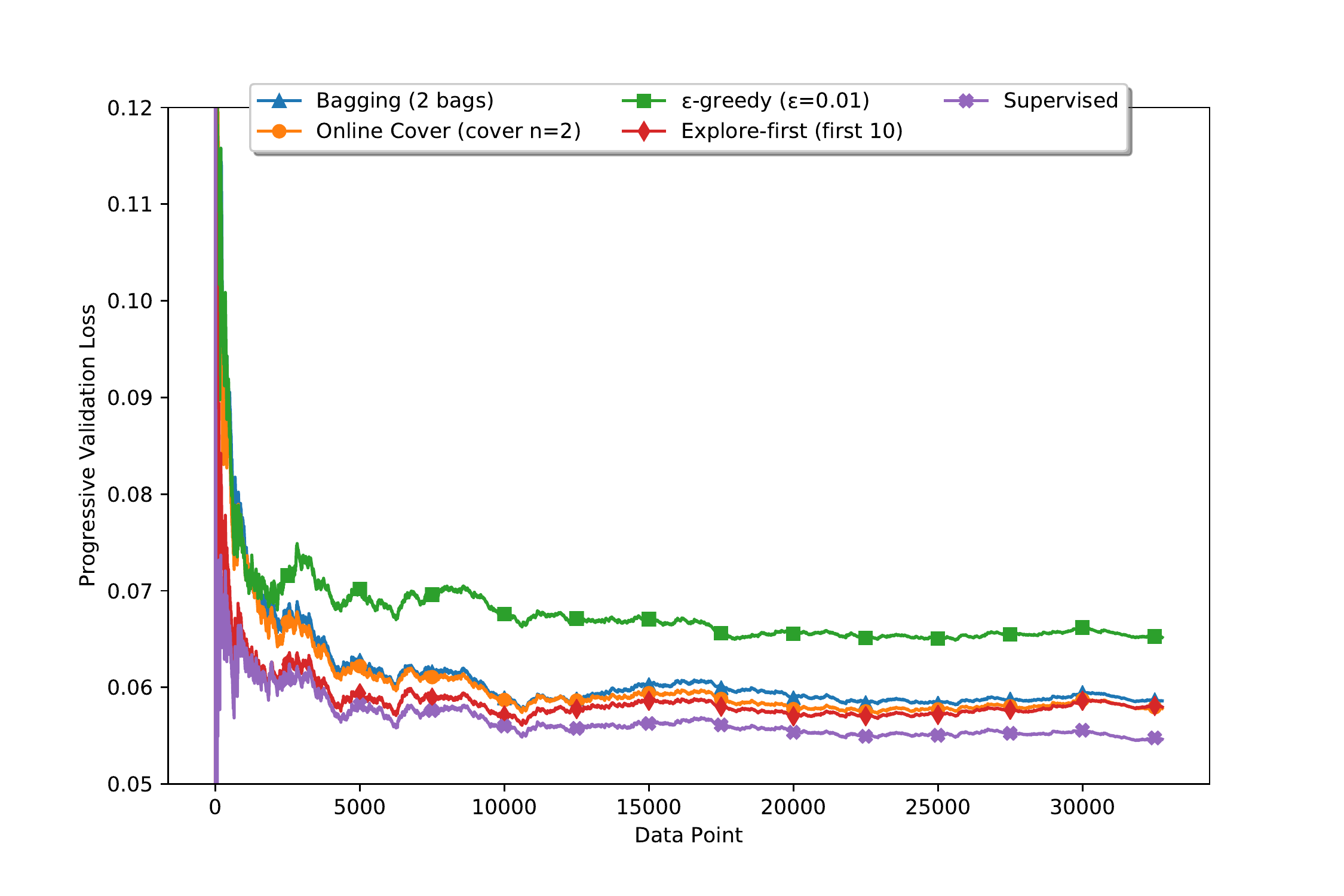}
    \caption{Progressive validation loss of various algorithms on Kaggle dataset \cite{kaggle}}
    \label{fig:kaggle}
\end{figure}

The results of different contextual bandit algorithms as well as a supervised classifier over the Kaggle dataset \cite{kaggle} are shown in Fig \ref{fig:kaggle} and reported in Table \ref{table:comparison}. The graph shows the progressive validation loss (P.V.Loss) \cite{blum1999beating} for each algorithm. We can see that under full information and by using a supervised algorithm, we get a pretty good predictor with an average loss rate of 5.5\%. Impressively, all contextual bandit algorithms get comparable performance on this dataset. Specifically, both Online Cover (with a cover set of size 2) and Explore-first (with first 10 records) algorithms get an average loss rate of 5.8\% which is very impressive considering the fact that compared to the full information supervised learning scenario they only have access to partial information.

\subsubsection{Manual and Synthetic Policy Datasets}
\begin{figure*}
        \centering
        \begin{subfigure}[b]{0.47\textwidth}
            \centering
            \includegraphics[scale=0.4]{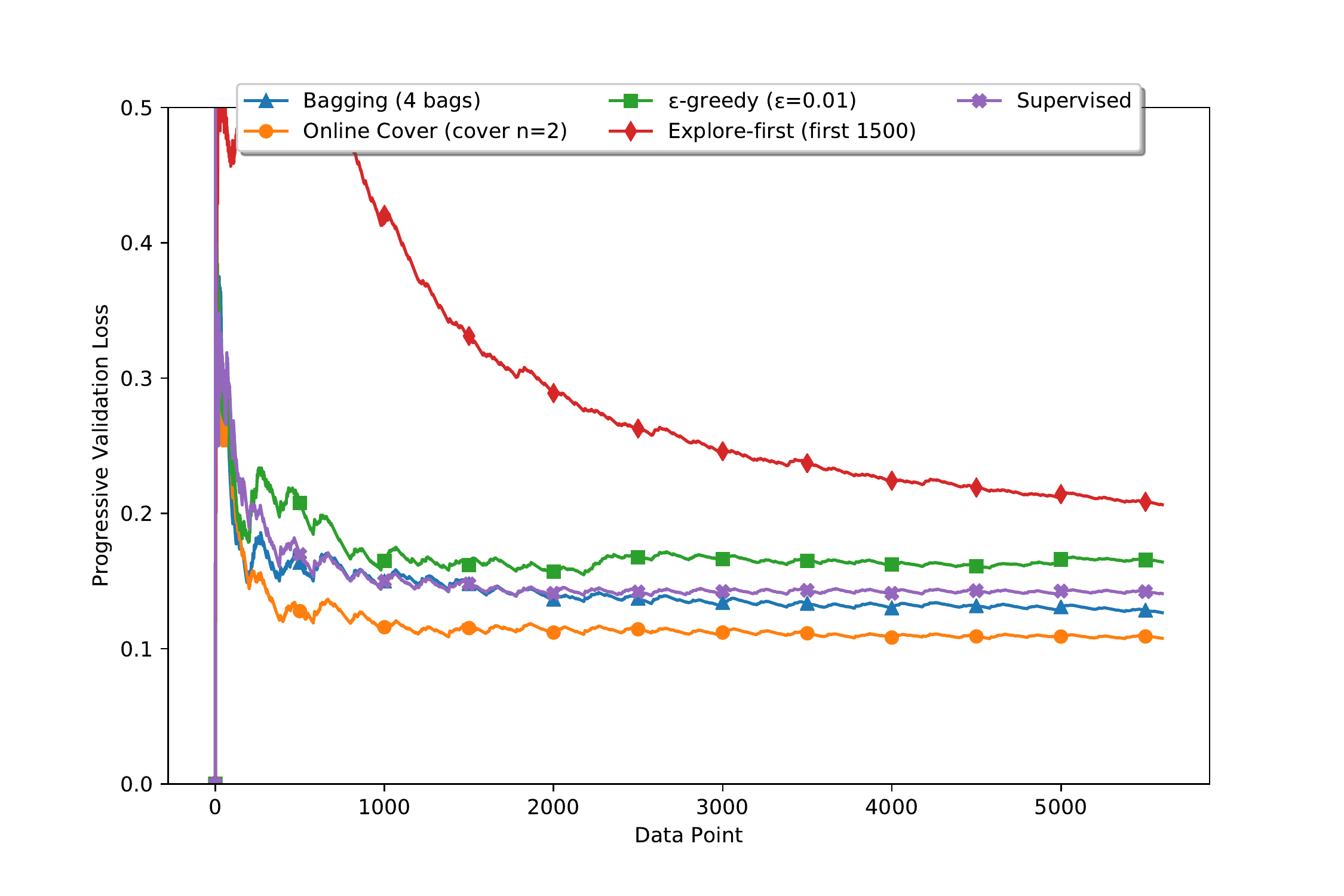}
            \caption[]%
            {{\small $\pi_{m1}$ Complete Dataset}} 
            \label{fig:rulem1_plot}
        \end{subfigure}
        \hfill
        \begin{subfigure}[b]{0.47\textwidth}  
            \centering 
            \includegraphics[scale=0.4]{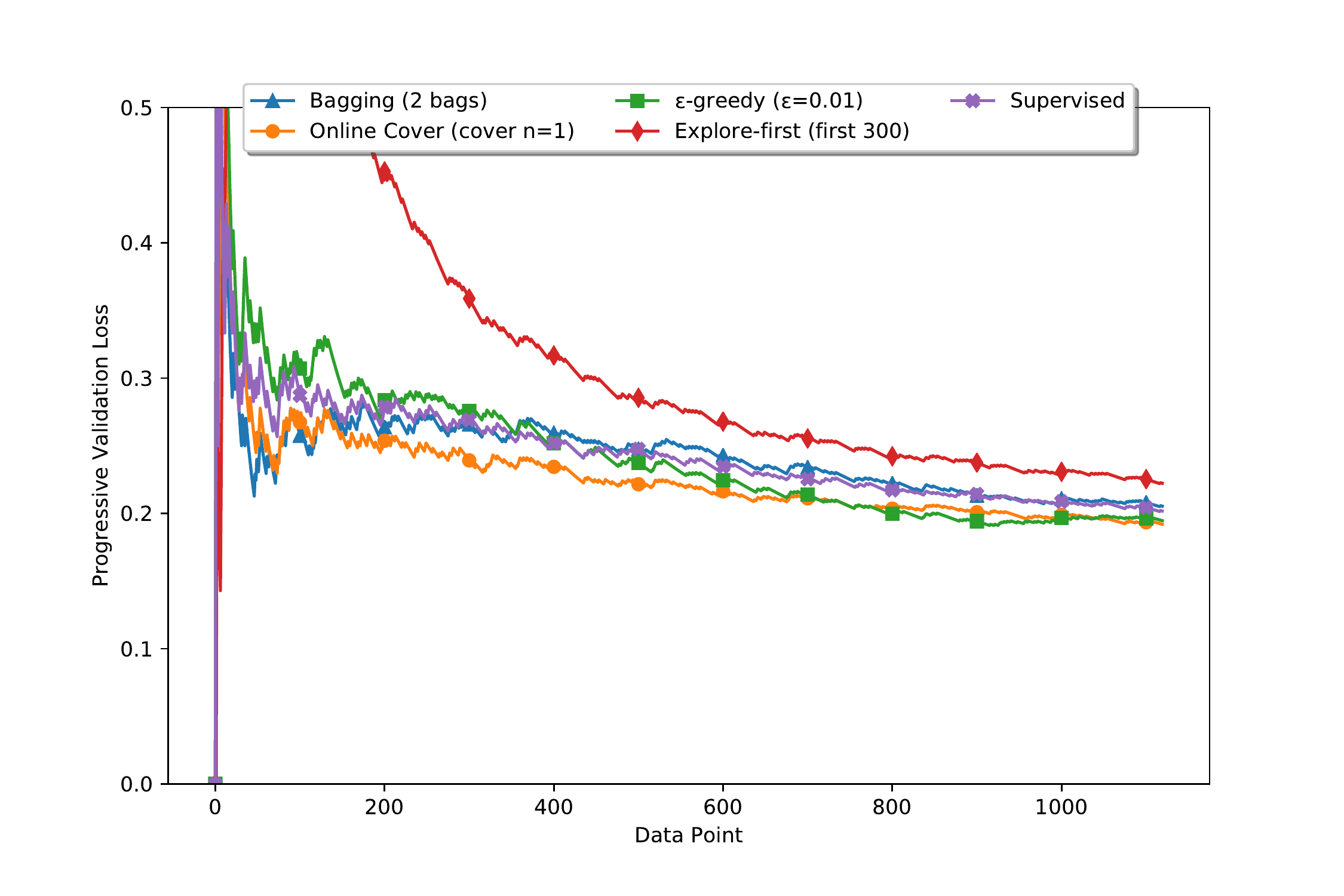}
            \caption[]%
            {{\small $\pi_{m1}$ Partial Dataset}} 
            \label{fig:rulem1_incomplete}
        \end{subfigure}
        \vskip\baselineskip
        \begin{subfigure}[b]{0.47\textwidth}   
            \centering 
            \includegraphics[scale=0.4]{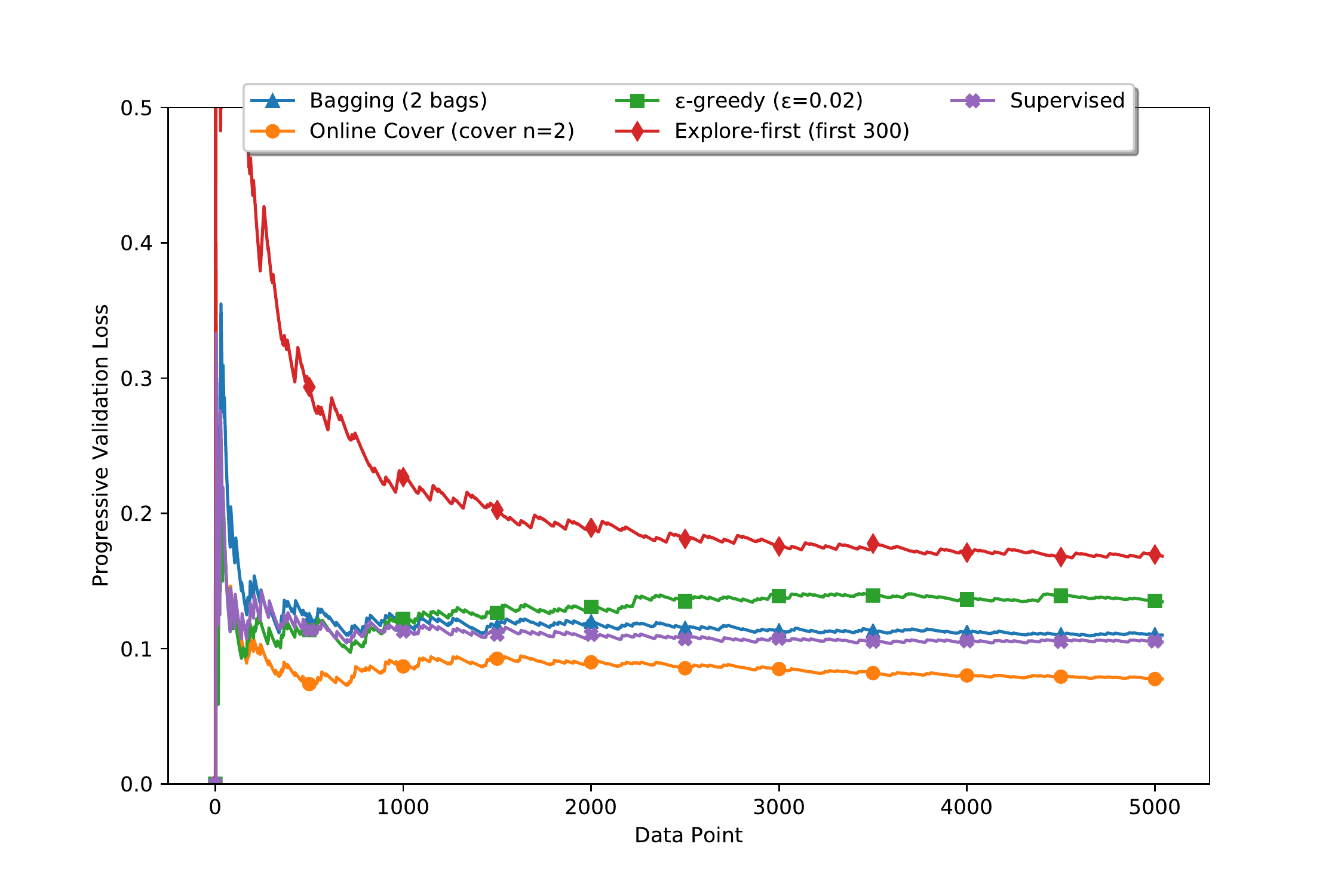}
            \caption[]%
            {{\small  $\pi_{m2}$ Complete Dataset}}    
            \label{fig:rulem2_plot}
        \end{subfigure}
        \hfill
        \begin{subfigure}[b]{0.47\textwidth}   
            \centering 
            \includegraphics[scale=0.4]{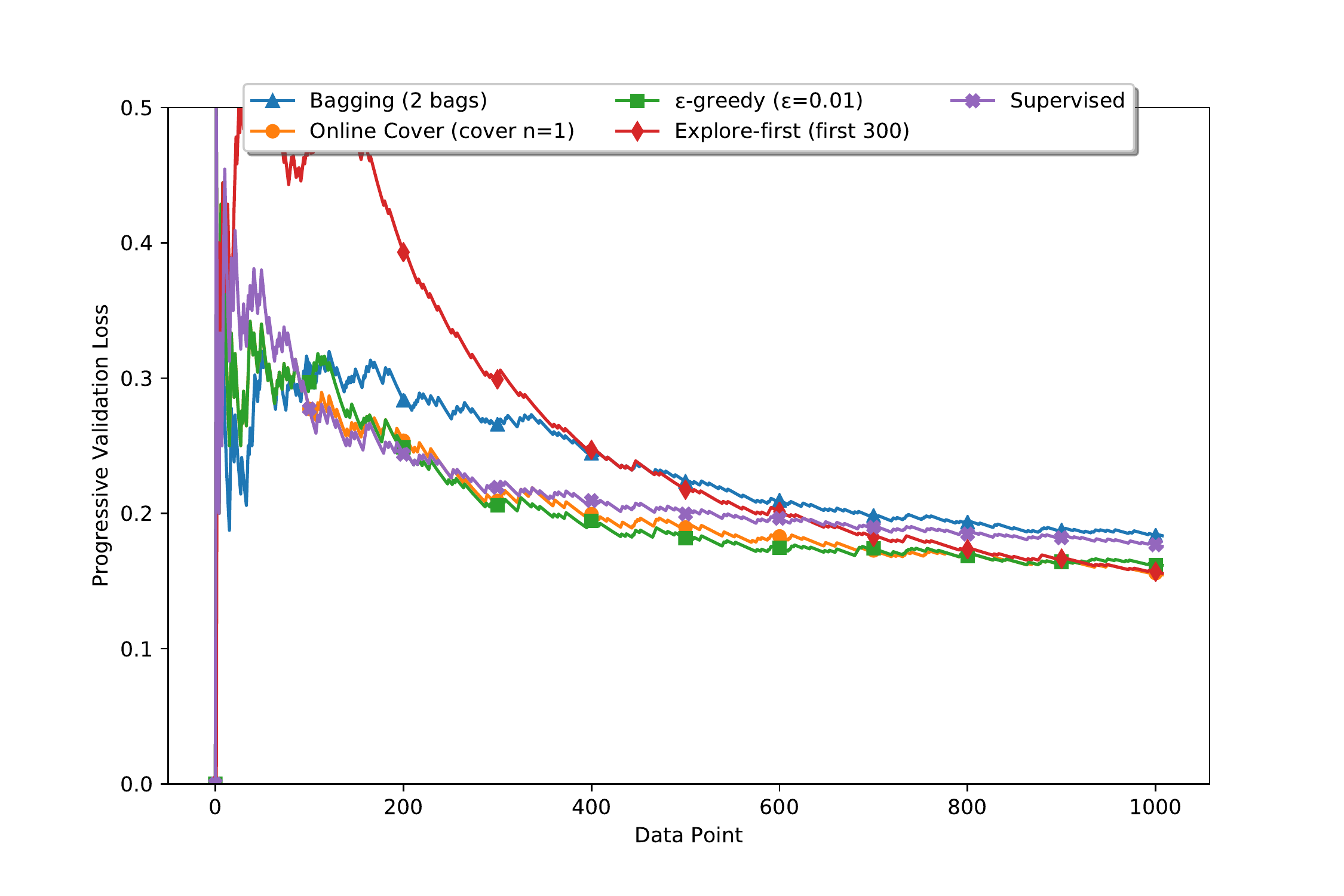}
            \caption[]%
            {{\small  $\pi_{m2}$ Partial Dataset}}    
            \label{fig:rulem2_incomplete}
        \end{subfigure}
        \vskip\baselineskip
        \begin{subfigure}[b]{0.47\textwidth}   
            \centering 
            \includegraphics[scale=0.4]{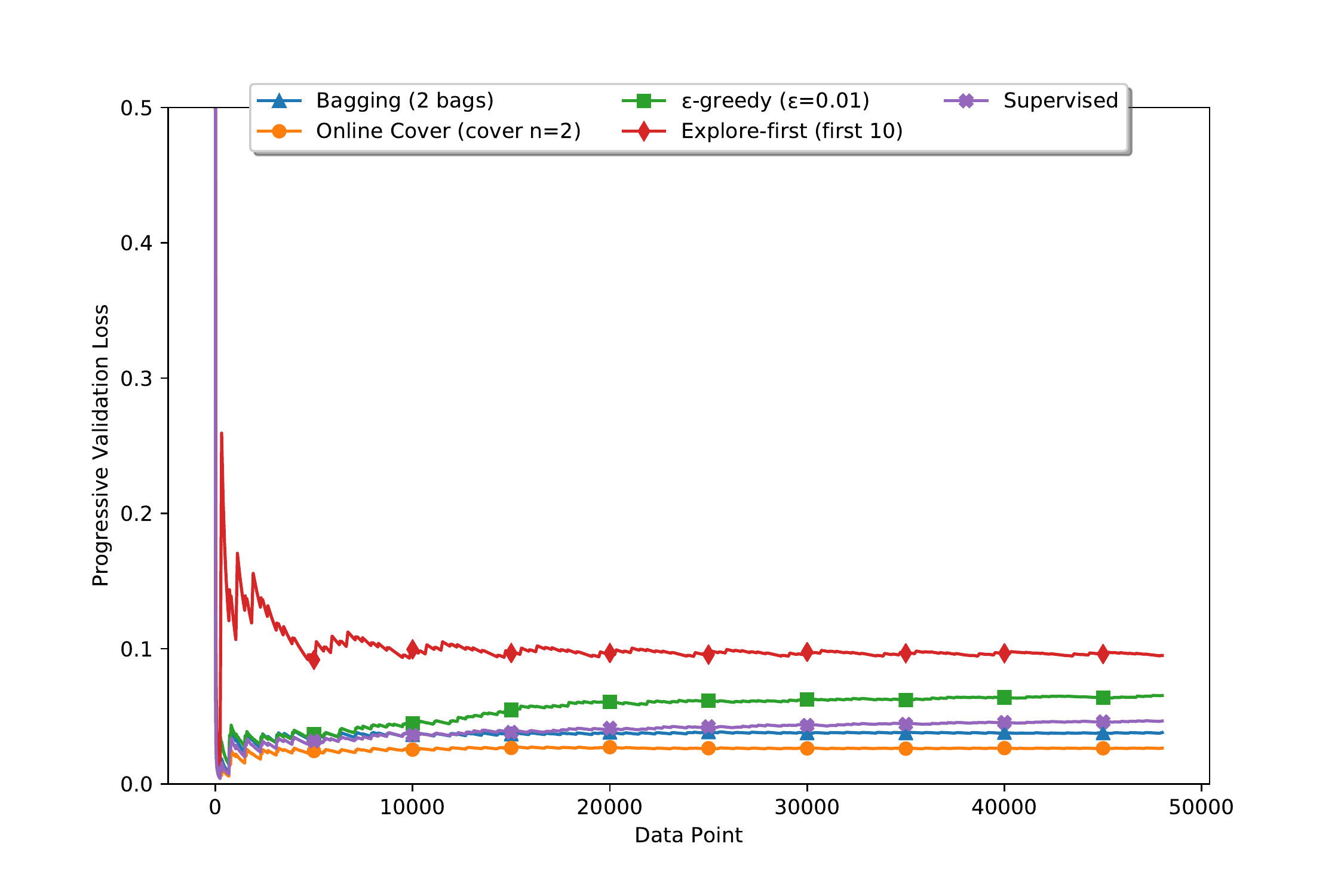}
            \caption[]%
            {{\small  $\pi_{m3}$ Complete Dataset}}    
            \label{fig:rulem3_plot}
        \end{subfigure}
        \hfill
        \begin{subfigure}[b]{0.47\textwidth}   
            \centering 
            \includegraphics[scale=0.4]{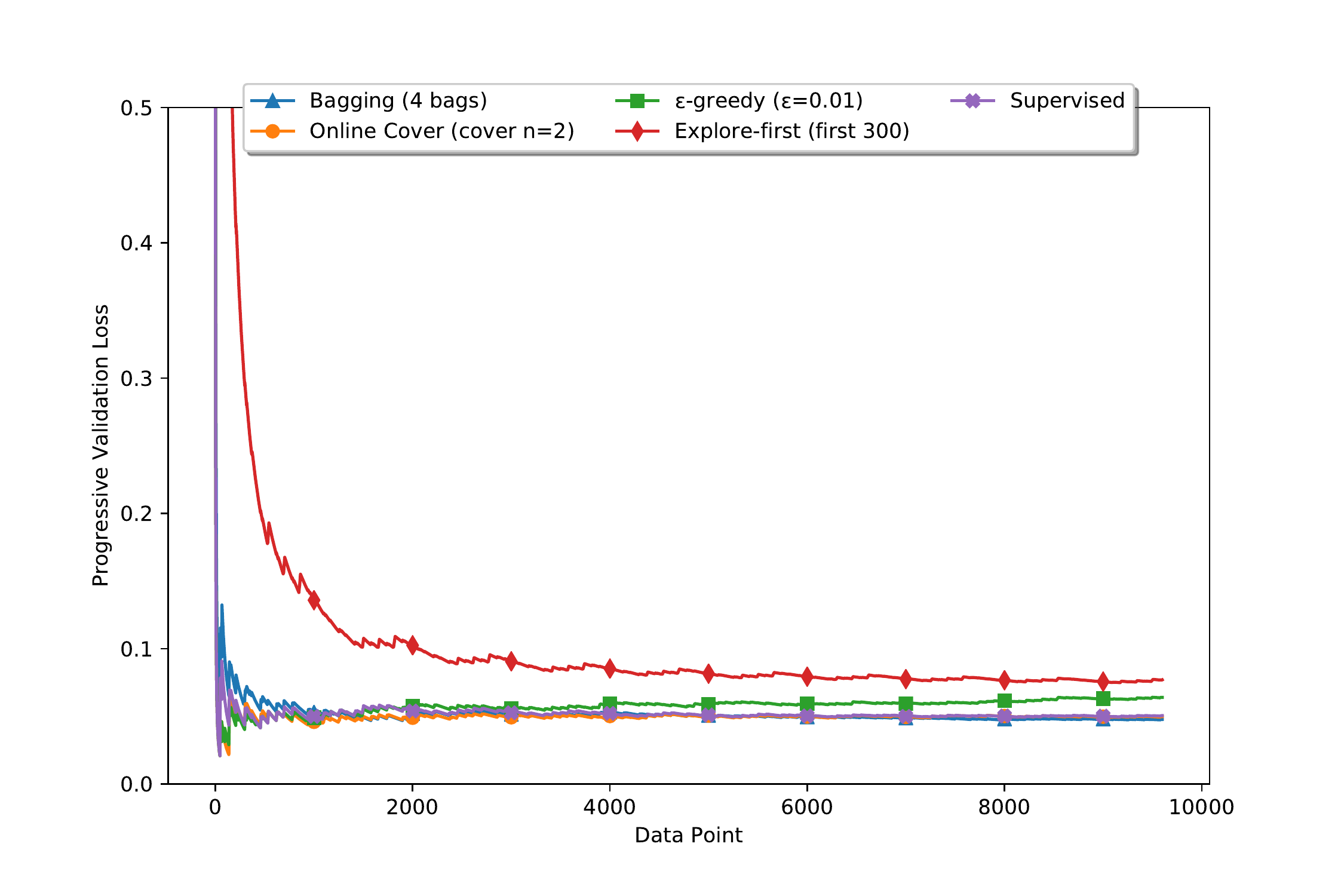}
            \caption[]%
            {{\small  $\pi_{m3}$ Partial Dataset}}    
            \label{fig:rulem3_incomplete}
        \end{subfigure}
        \caption[]
        {\small Progressive validation loss of various algorithm on manual policies' complete and partial datasets} 
        \label{fig:manual_policies}
\end{figure*}

\begin{figure*}
        \centering
        \begin{subfigure}[b]{0.47\textwidth}
            \centering
            \includegraphics[scale=0.4]{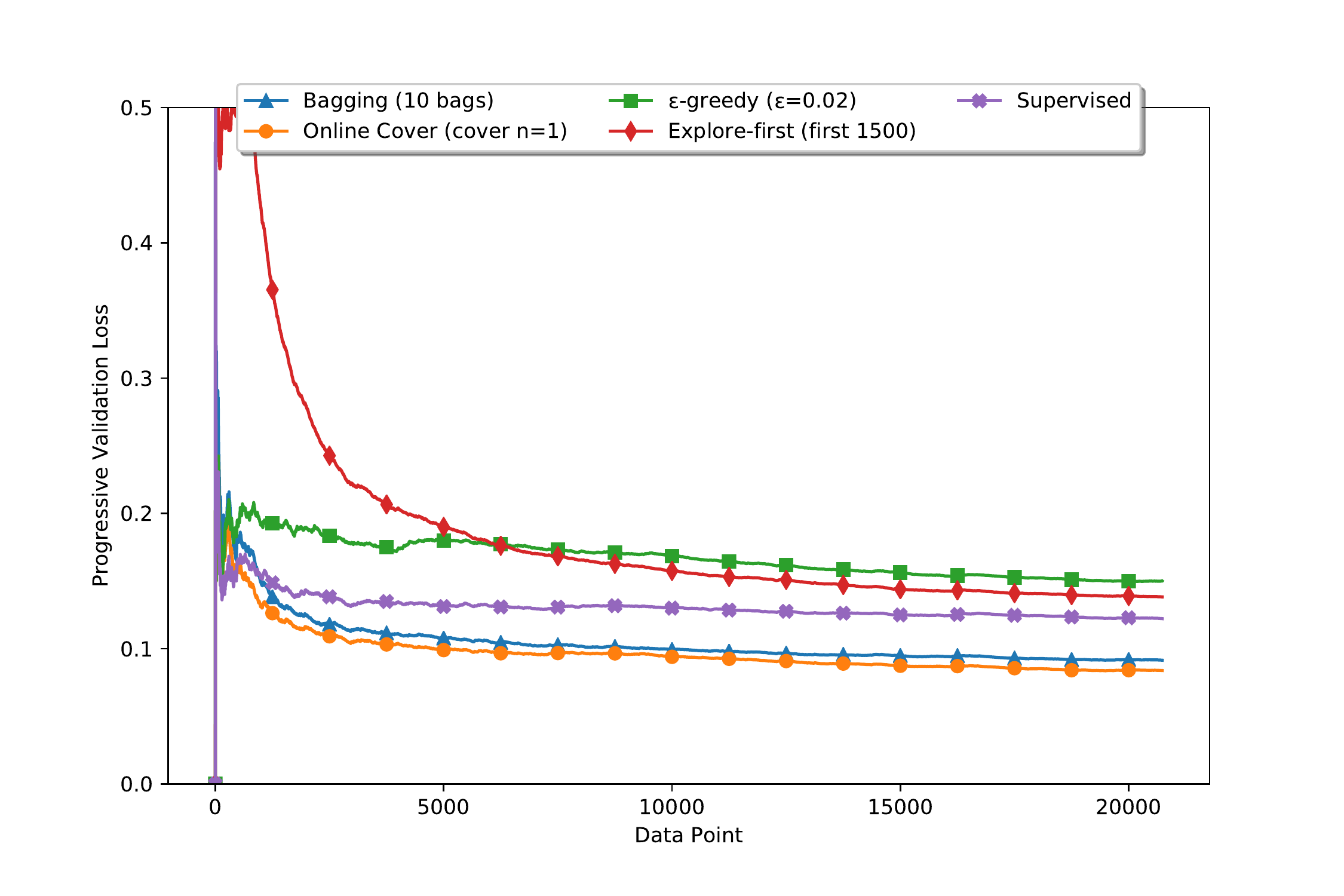}
            \caption[]%
            {{\small $\pi_{s1}$ Complete Dataset}} 
            \label{fig:rules1_plot}
        \end{subfigure}
        \hfill
        \begin{subfigure}[b]{0.47\textwidth}  
            \centering 
            \includegraphics[scale=0.4]{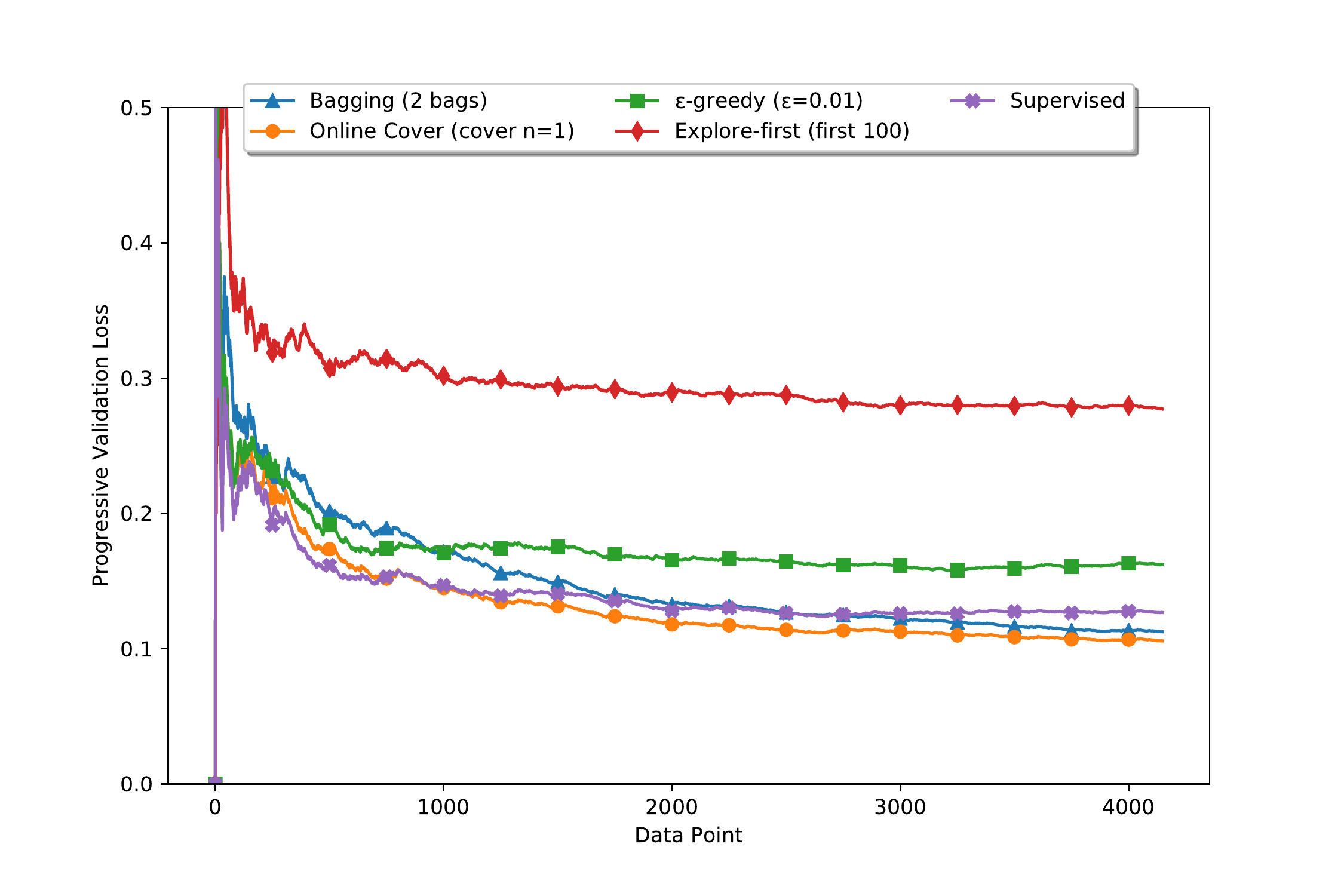}
            \caption[]%
            {{\small $\pi_{s1}$ Partial Dataset}} 
            \label{fig:rules1_incomplete}
        \end{subfigure}
        \vskip\baselineskip
        \begin{subfigure}[b]{0.47\textwidth}   
            \centering 
            \includegraphics[scale=0.4]{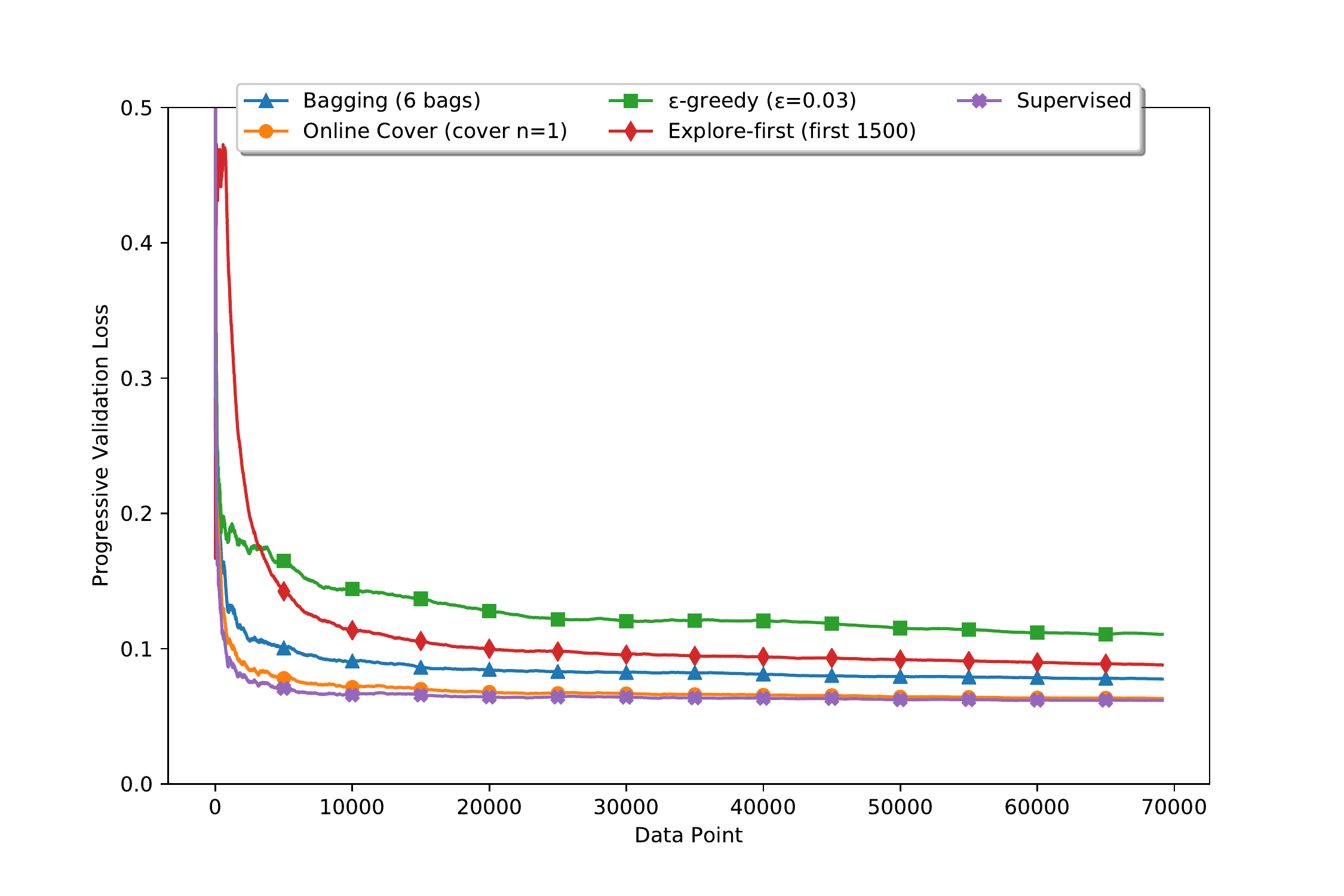}
            \caption[]%
            {{\small  $\pi_{s2}$ Complete Dataset}}    
            \label{fig:mean and std of net34}
        \end{subfigure}
        \hfill
        \begin{subfigure}[b]{0.47\textwidth}   
            \centering 
            \includegraphics[scale=0.4]{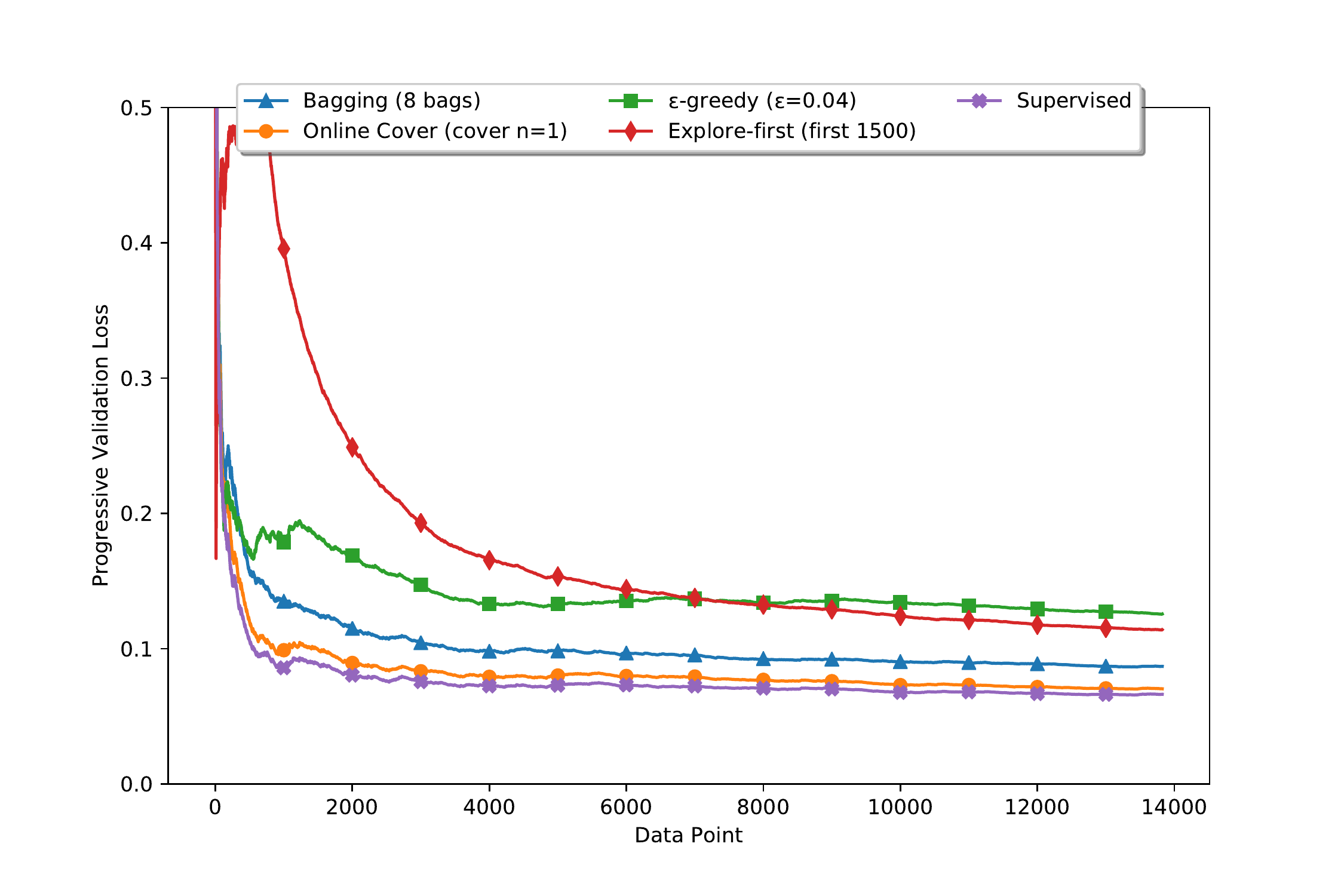}
            \caption[]%
            {{\small  $\pi_{s2}$ Partial Dataset}}    
            \label{fig:mean and std of net44}
        \end{subfigure}
        \vskip\baselineskip
        \begin{subfigure}[b]{0.47\textwidth}   
            \centering 
            \includegraphics[scale=0.4]{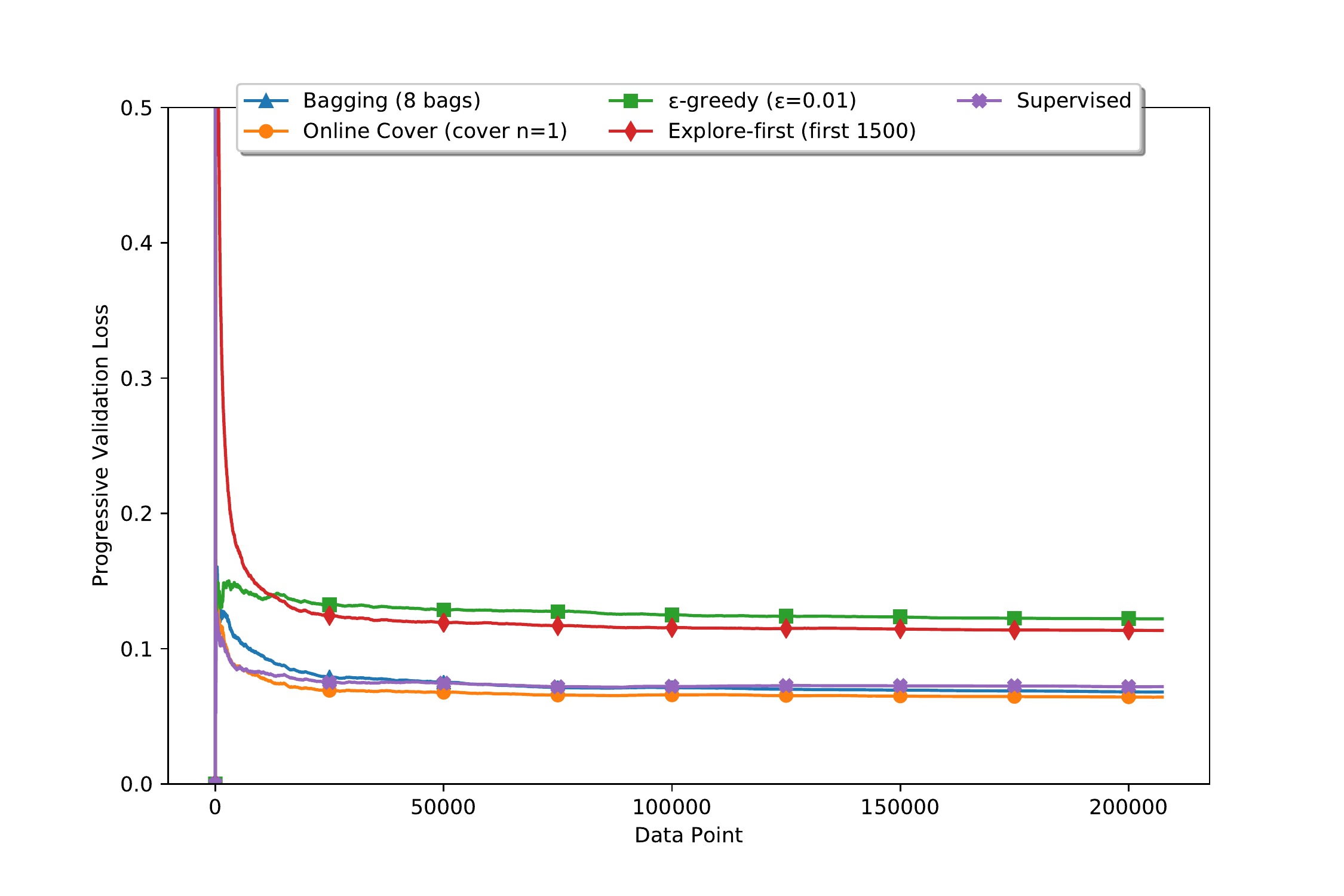}
            \caption[]%
            {{\small  $\pi_{s3}$ Complete Dataset}}    
            \label{fig:mean and std of net34}
        \end{subfigure}
        \hfill
        \begin{subfigure}[b]{0.47\textwidth}   
            \centering 
            \includegraphics[scale=0.4]{images/rule3_incomplete_plot.pdf}
            \caption[]%
            {{\small  $\pi_{s3}$ Partial Dataset}}    
            \label{fig:mean and std of net44}
        \end{subfigure}
        \caption[]
        {\small Progressive validation loss of various algorithm on synthetic policies' complete and partial datasets} 
        \label{fig:synthetic_policies}
\end{figure*}

Figures \ref{fig:manual_policies} and \ref{fig:synthetic_policies} show the results of various learning algorithms over complete and partial datasets for manually written policies as well as randomly generated ones. Interestingly, in most cases, one or more contextual bandit algorithms outperform the supervised learning one. For all complete datasets, the online cover algorithm converges to the lowest validation loss. As we expect, algorithms achieve lower loss over complete datasets compared to the partial ones as they have more data available in their training phase. In the same way, as it is shown in Fig. \ref{fig:synthetic_policies}, as datasets get larger, final losses of algorithms over them become lower.  

\begin{figure}[htbp]
    \centering\includegraphics[scale=0.4]{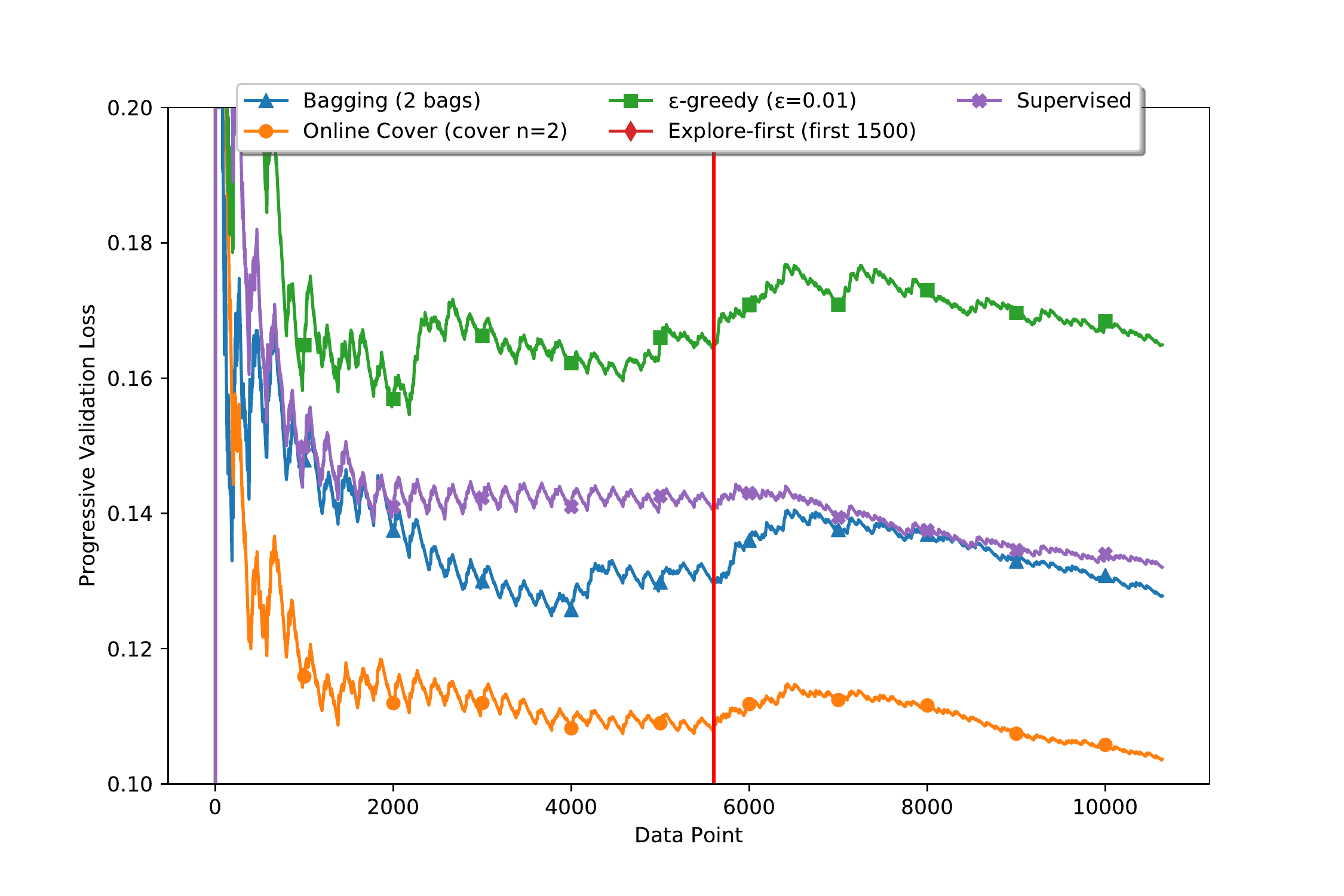}
    \caption{Adaptation of various algorithms to policy shift}
    \label{fig:policy_shift}
\end{figure}

\subsubsection{Policy Shift}\label{policy_shift} Next, we examine how various algorithms respond to a shift in the authorization policy. Fig \ref{fig:policy_shift} shows the P.V.Loss of these algorithms over a policy that has a shift after $t=5600$. The authorization policy shifts from $\pi_{m1}$ to $\pi_{m2}$ after this timestamp. Note that all algorithms have a slight rise in P.V.Loss after the shift but they adapt to the change in the policy. For this specific simulation, Online Cover (with a cover set of size 2) adjusts quicker than other algorithms and converge to a better P.V.Loss value. Interestingly, the two contextual bandit algorithms (i.e. Online Cover and Bag) outperform the supervised learning algorithm in terms of adapting to the shift in the policy. 

\subsubsection{Policy Initialization Techniques}
\begin{figure}[htbp]
    \centering\includegraphics[scale=0.5]{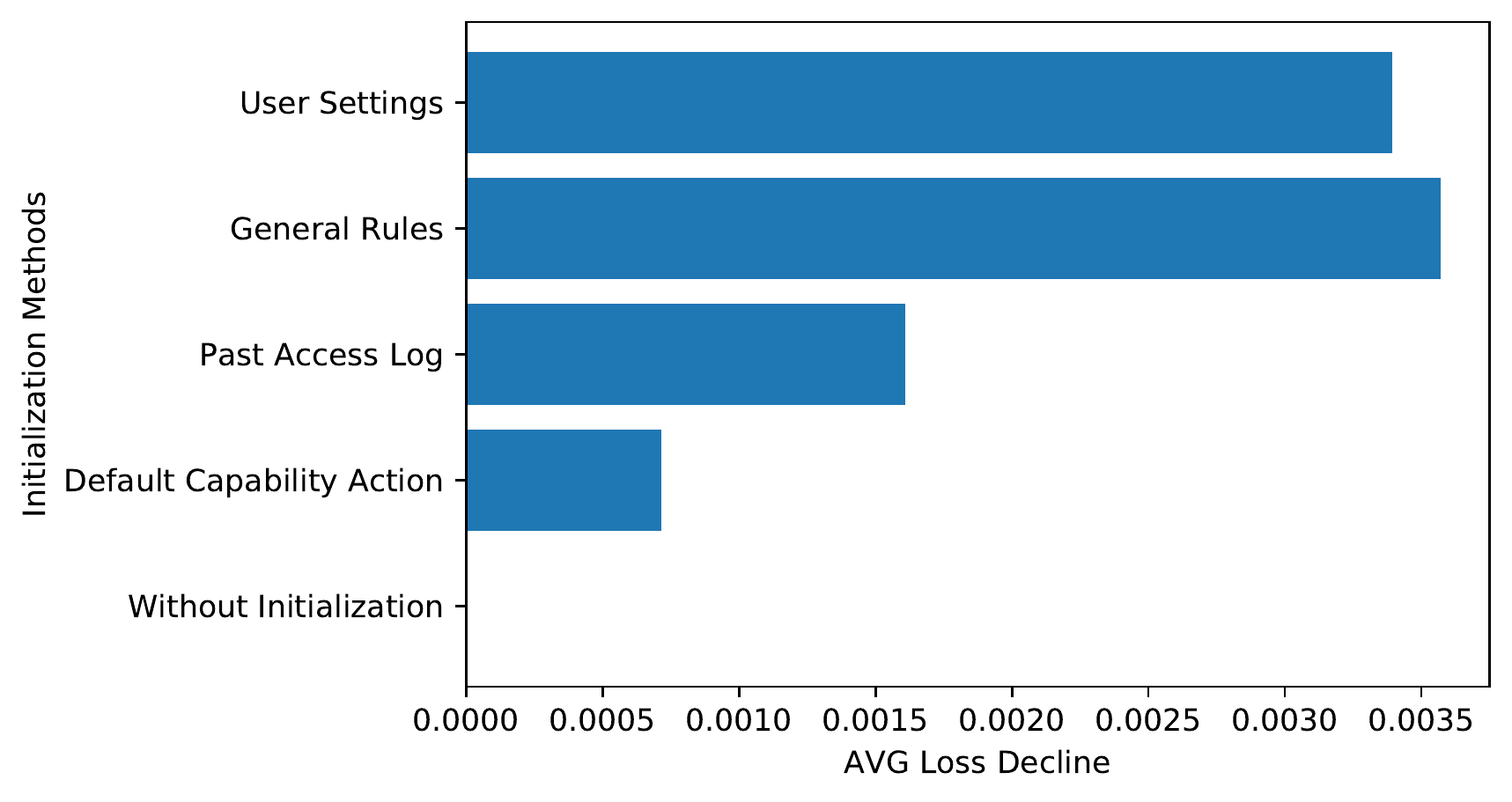}
    \caption{The effect of various initialization techniques in reducing average loss of Online Cover model on $\pi_{m3}$ dataset}
    \label{fig:initialization}
\end{figure}

Fig \ref{fig:initialization} shows the results of various initialization techniques over manual policy $\pi_{m3}$ dataset. As shown in the graph, the initialization with general rules resulted in the highest decline in the average loss and the initialization based on the default capability action resulted in the lowest decline in the average loss. 

\subsubsection{Policy Learning with Planning}
\begin{figure}[htbp]
    \centering\includegraphics[scale=0.4]{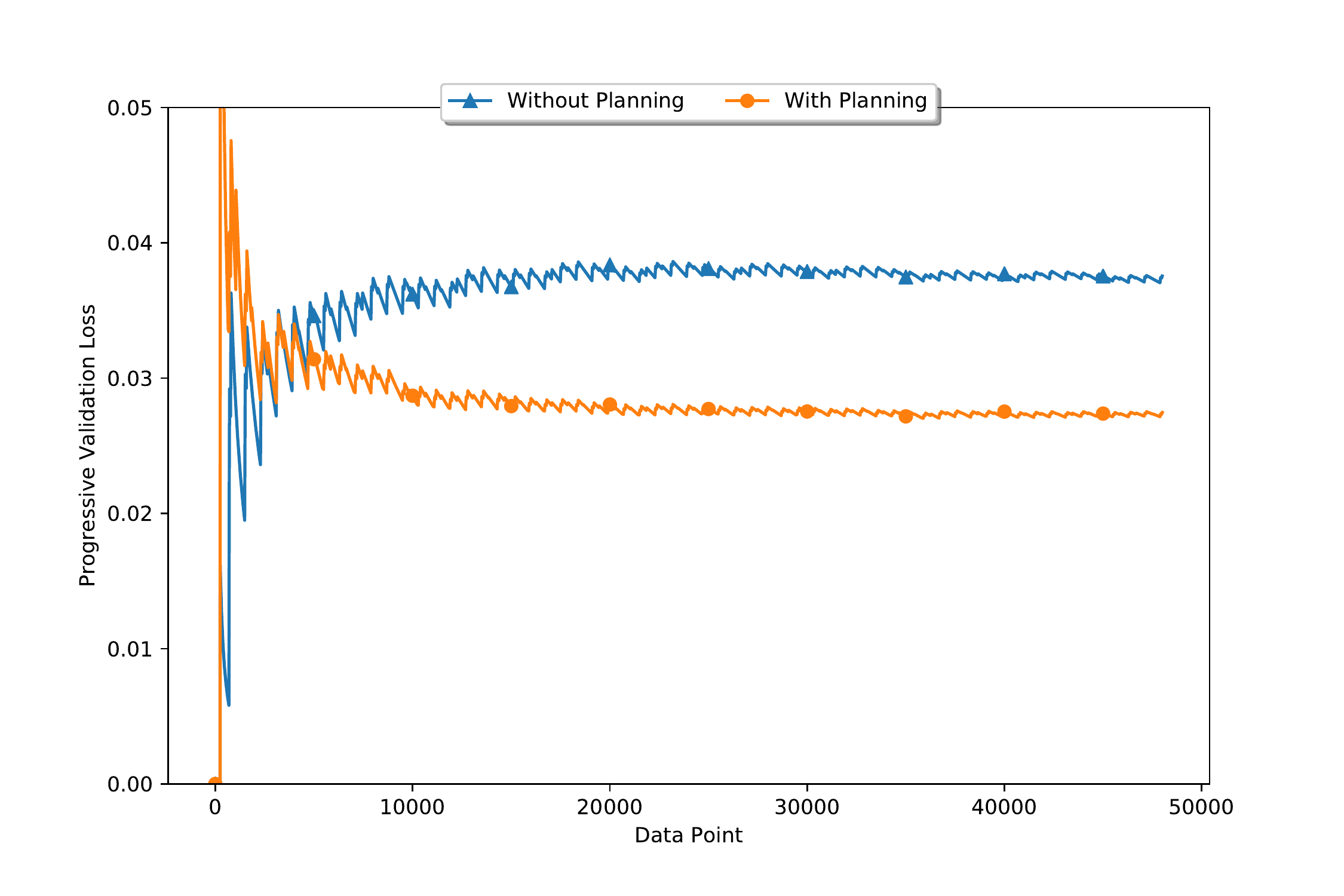}
    \caption{The effect of learning with planning on reducing average loss of Online Cover model on $\pi_{m3}$ dataset}
    \label{fig:planning}
\end{figure}

Fig \ref{fig:planning} shows the results of the Online Cover algorithm over manual policy $\pi_{m3}$ with and without planning. As we can see in the figure, the learning algorithm with planning decreased the PVL by 25\% for this dataset. We should note that Online Cover learned a model with the lowest PVL for this dataset and this is a significant reduction for such a model. Table \ref{table:comparison} shows the results of the planning algorithm over various databases. As we can see from the results, the planning algorithm decreases the PVL for at least 10\% for all cases that planning was applied. 

\section{Related Work}\label{relatedwork}
In this section, we discuss related research areas. We cover the application of Reinforcement Learning in real-world problems, adaptive authorization models, and finally access control in the IoT environment. 

\subsection{Application of Reinforcement Learning}
Reinforcement learning has been applied to a variety of real-world problems where access to pre-labeled data is limited. Around three decades ago, Boyan and Littman employed RL for packet routing in switches \cite{boyan1994packet}. Recently, RL is used in designing protocols for controlling congestion \cite{dong2015pcc} as well as resource management in networks \cite{mao2016resource} and resource allocation for vehicle-to-vehicle communications \cite{ye2019deep}. Furthermore, Contextual bandit algorithms are being employed for solving many real-world interactive machine learning problems.

% Deep reinforcement learning was very successful in achieving human-level proficiency in playing video games for both 2D \cite{mnih2013playing} and 3D \cite{lample2017playing} environments in which full knowledge of the current state of the environment is not available and states are partially observable. 

Although RL has proved to be a promising solution in many domains, we are not aware of any work that applies reinforcement learning to access control policy adaption and management.

\subsection{Adaptive Authorization Models}
The manual development of authorization systems is time-consuming and error-prone. Furthermore, policy misconfigurations reduce the effectiveness of access control systems. Multiple studies try to address such challenges. RBAC policy learning models try to find the optimal set of roles from user permission \cite{molloy2010mining, xu2012algorithms, kuhlmann2003role, mitra2016survey}. Subsequently, ABAC policy learning models are proposed to ease the migration to ABAC systems and development of ABAC policies \cite{xu2015mining, cotrini2018mining, karimi2018unsupervised, iyer2018mining, karimi2020automatic}. These models extract ABAC policy rules from available access logs to automate the development of an authorization system.

Furthermore, there has been work on adaptive access control models in which the access control models are adjusted based on users' activity. Baracaldo and Joshi \cite{baracaldo2013adaptive} incorporate risk and trust assessment in RBAC so the framework adapts to suspicious shifts in users' behavioral patterns by removing privileges when users' trust score is lower than a threshold. Marinescu \textit{et al.} \cite{marinescu2017ivd} propose an approach for detecting authorization bugs in an online social network system and blocking access attempts that try to exploit such bugs. Their proposed approach learns authorization
rules from data manipulation patterns and enforces such rules to prevent unauthorized accesses before code fixes are deployed.
Argento \textit{et al.} \cite{argento2018towards} propose an adaptive access control model that exploits users' behavioral patterns to narrow their permissions when anomalous behavior is detected. 

\subsection{Access Control in IoT Environment}
Although we have seen a proliferation of IoT devices in recent years, commercial IoT infrastructures suffer from proper access control mechanisms. Recently, few studies have addressed authorization specification and enforcement in IoT environments.
Fernandes \textit{et al.} discuss how coarse-grained capabilities have resulted in over-privileged smart home applications \cite{fernandes2016security}. Several works try to mitigate the challenge by rethinking permission granting \cite{fernandes2016security, jia2017contexlot, tian2017smartauth}. Ur \textit{et al.} have examined access control affordances in smart devices and discussed how they are insufficient from a usability perspective \cite{ur2013current}.

He \textit{et al.} \cite{he2018rethinking} propose a capability-centric access control specification for the home IoT devices which considers capabilities of devices, relationships between users, and some contextual information in the authorization policy specification. Tian \textit{et al.} \cite{tian2017smartauth} present \textit{SmartAuth} a user-centric and semantic-based authorization framework, which collects security-relevant information from the description, code and annotations of IoT applications to guide users to make well-informed decisions when authorizing IoT applications. Fernandes \textit{et al.} \cite{197137} propose FlowFence, which requires the consumer applications of sensitive data to declare their intended data flow. FlowFence blocks any flow that is not declared.

\section{Discussion}\label{discussion}
While an adaptive ABAC policy learning through an RL model seems a promising approach for policy development and management in complex and evolving systems, there are some considerations that need to be reviewed carefully. In this section, we discuss such challenges.

\subsection{Managing Incorrect Authorization Decisions}
The learning process need to go through multiple rounds of state-action-feedback sequences to achieve good performance. Hence, some incorrect authorization decisions during the learning phase is inevitable. To address the issue, the system could employ the following approaches to minimize the consequences of the wrong authorization decisions of the RL agent throughout the learning process:

\begin{itemize}
    \item The system can employ a simpler AC model (e.g., RBAC or DAC) while the learning process is happening.
    \item The system can continue using its legacy AC model while the learning process is happening and before completely migrating to the newly learned ABAC model. This approach has two advantages: first, the decision of the legacy system can be used as feedback to the learning agent, and second, the migration can happen when the administrator feels confident about the decisions of the agent (i.e., the loss is less than a desirable threshold).
    \item For sensitive resources, the default action can be set to ``deny” (except for an administrator) and the RL model can be initialized with these default decisions as suggested in the paper.
\end{itemize}

\subsection{Model Convergence}
The learning process of the proposed model is assumed to have been converged when the loss of the RL model is less than a desired threshold set by the administrator of the system. However, in the case of dynamic systems, for any newly added attribute/attribute values or an update in authorization policies, a learning process will continue. As such, the RL agent will always work in the background.

For example, as we can see in Figure \ref{fig:manual_policies} and \ref{fig:synthetic_policies}, the more the agent receives feedback from the users, the lower the loss of the RL model would be. If we assume the desired loss of the model is set at 0.15 by the administrator of the system, then the Online Cover algorithm converges for all the datasets. 

On the other hand, when we have an update in the authorization policy (as described in Section \ref{policy_shift} and was shown in Figure \ref{fig:policy_shift}), although the RL model was converged before the policy update, the loss rate would increase after such update. Hence, the agent needs more feedback from the users to learn the new policy. Therefore, the loss rate decreases and the model converges again. 

\section{Conclusion}\label{conclusion}
In this paper, we take a reinforcement learning, more specifically, a contextual-bandit approach, to support adaptive ABAC policy learning. We have proposed a simple and reliable method for learning ABAC policy rules from access logs by incorporating the feedback of users on access decisions made by the authorization engine. We have focused on Home IoT as a running example throughout the paper. However, the proposed model can be applied to any domain as long as it is not highly sensitive (which is decided by the administrator of the system) or unless the loss threshold is set too low before the model is employed in the real-world. Especially, in the environments where the usability of the model is a priority, the proposed framework helps lifting the burden of developing authorization policies from incompetent users of the system.

In addition, we have proposed four different procedures for initializing the learning model and a planning approach based on attribute value hierarchies to accelerate the learning process. We have evaluated our proposed approach over real and synthetic data including both complete and sparse datasets. Our experiments show that the proposed learning model achieves comparable performance to the full information supervised learning methods in many scenarios and even outperforms them in several situations. 

% %-------------------------------------------------------------------------------
% \section*{Acknowledgments}
% %-------------------------------------------------------------------------------

% The USENIX latex style is old and very tired, which is why
% there's no \textbackslash{}acks command for you to use when
% acknowledging. Sorry.

% %-------------------------------------------------------------------------------
% \section*{Availability}
% %-------------------------------------------------------------------------------

% USENIX program committees give extra points to submissions that are
% backed by artifacts that are publicly available. If you made your code
% or data available, it's worth mentioning this fact in a dedicated
% section.

%-------------------------------------------------------------------------------
\bibliographystyle{plain}
\bibliography{bibliography}

\clearpage
\appendix
\section{Sample Manual Policies} \label{sample}
In the following, we present details of the manual policies we used in our experiments. We defined three sample manual policies for our experiments. Table \ref{table:manual_policies_op} and Table \ref{table:manual_policies_attr} show their operations, attributes, and corresponding attribute values. 

\begin{table}[!htbp]
\small
\centering
\caption{Operations in sample manual policies}
\begin{tabular}{ll}
    \toprule
    Policy & Operation\\
    \midrule
    \multirow{10}{*}{Manual Policy 1 ($\pi_{m1}$)} & lights\_on\_off \\ & order\_online \\ & set\_temperature \\ & turn\_on\_cooler \\ & turn\_on\_heater \\ & install\_software\_update \\ & mower\_on\_off \\ & connect\_new\_device \\ & view\_lock\_state \\ & play\_music \\
     \hline
     \multirow{9}{*}{Manual Policy 2 ($\pi_{m2}$)} & lights\_on\_off \\ & order\_online \\ & set\_temperature \\ & play\_music \\ & turn\_on\_cooler \\ & turn\_on\_heater \\ & camera\_on\_off \\ & view\_temperature\_log \\ & answer\_door \\
     \hline
     \multirow{10}{*}{Manual Policy 3 ($\pi_{m3}$)} & lights\_on\_off \\ & order\_online \\ & set\_temperature \\ & play\_music \\ & turn\_on\_cooler \\ & turn\_on\_heater \\ & camera\_on\_off \\ & view\_temperature\_log \\ & answer\_door \\ & mower\_on\_off \\
    \bottomrule
    \end{tabular}
\label{table:manual_policies_op}
\end{table}

\begin{table}
\small
\centering
\caption{Attributes and their corresponding values of sample manual policies}
\begin{tabular}{lll}
    \toprule
    Policy & Attribute &  Attribute Value \\
    \midrule
    \multirow{13}{*}{Manual Policy 1 ($\pi_{m1}$)} & \multirow{4}{*}{Time} & day \\ & & midday \\ & & night \\ & & midnight \\
    \cline{2-3}
      & \multirow{5}{*}{Role} & mother \\ & & father \\ & & child \\ & & visiting\_family \\ & & guest \\
     \cline{2-3}
     & \multirow{4}{*}{Location} & inside\_home \\ & & outside\_home \\ & & yard \\ & & basement \\
    % %  & Username & M, F, D, S, B, N, BS\\
    % %  & Operation & lights\_on\_off, order\_online, set\_temperature,turn\_on\_cooler, \\ 
    % %  & & turn\_on\_heater, install\_software\_update, mower\_on\_off, \\
    % %  & & connect\_new\_device, view\_lock\_state, play\_music \\
     \hline
     \multirow{13}{*}{Manual Policy 2 ($\pi_{m2}$)} & \multirow{4}{*}{Time} & morning \\ & & afternoon \\ & & evening \\ & & night \\
     \cline{2-3}
      & \multirow{5}{*}{Role} & mother \\ & & father \\ & & child \\ & & baby\_sitter \\ & & neighbor \\
     \cline{2-3}
      & \multirow{4}{*}{Location} & kitchen \\ & & living\_room \\ & & bedroom1 \\ & & bedroom2 \\
    % %   & Username & M, F, D, S, B, N, BS\\
    % %   & Operation & "lights\_on\_off", "order\_online", "set\_temperature", "play\_music",   \\
    % %   & & "turn\_on\_cooler", "turn\_on\_heater", "camera\_on\_off", \\
    % %   & & "view\_temperature\_log", "answer\_door" \\
     \hline
     \multirow{14}{*}{Manual Policy 3 ($\pi_{m3}$)} & \multirow{6}{*}{Time} & day \\ & & morning \\ & & afternoon \\ & & evening \\ & & night \\ & & midnight \\
     \cline{2-3}
     & \multirow{8}{*}{Location} & kitchen \\ & & living\_room \\ & & bedroom1 \\ & & bedroom2 \\ & & inside\_home \\ & & outside\_home \\ & & yard \\ & & basement \\
     \cline{2-3}
     & \multirow{10}{*}{Role} & parent \\ & & mother \\ & & father \\ & & child \\ & & minor\_child \\ & & teenager \\ & & guest \\ & & baby\_sitter \\ & & neighbor \\ & & visiting\_family \\
    %  & Username & M, F, D, S, B, N, BS, G, T, P \\
    %  & Operation & lights\_on\_off, order\_online, set\_temperature, play\_music, \\
    %  & & turn\_on\_cooler, turn\_on\_heater, camera\_on\_off, \\
    %  & & view\_temperature\_log, answer\_door, mower\_on\_off \\
    \bottomrule
    \end{tabular}
\label{table:manual_policies_attr}
\end{table}

%%%%%%%%%%%%%%%%%%%%%%%%%%%%%%%%%%%%%%%%%%%%%%%%%%%%%%%%%%%%%%%%%%%%%%%%%%%%%%%%
\end{document}